\documentclass[final,3p,times,twocolumn]{elsarticle}
\usepackage{amsmath,amsfonts}
\usepackage{algorithmic}
\usepackage{algorithm}
\usepackage{array}
\usepackage[caption=false,font=normalsize,labelfont=sf,textfont=sf]{subfig}\usepackage[table,xcdraw]{xcolor}
\usepackage{float}
\usepackage{textcomp}
\usepackage{stfloats}
\usepackage{url}
\usepackage{verbatim}
\usepackage{graphicx}

\usepackage{amssymb}
\usepackage{amsmath}
\journal{Nuclear Physics B}

\begin{document}

\begin{frontmatter}


%%       addressline={},
%%       city={},
%%       postcode={},
%%       state={},
%%       country={}}

\title{Distilled Large Language Model-Driven Dynamic Sparse Expert Activation  Mechanism}

%%       addressline={},
%%       city={},
%%       postcode={},
%%       state={},
%%       country={}}
%%       addressline={},
%%       city={},
%%       postcode={},
%%       state={},
%%       country={}}
\author[1,2]{Qinghui Chen\fnref{label1}}\author[1]{Zekai Zhang\fnref{label1}}\author[3]{Zaigui Zhang}\author[3]{Kai Zhang}\author[4]{Dagang Li}\author[4]{Wenmin Wang}\author[1]{Jinglin Zhang\corref{cor1}}\author[5]{Cong Liu}

% \author[1,2]{Qinghui Chen}\author[1]{Zekai Zhang}\author[2]{Hailong Liu}\author[1]{Jinglin Zhang\corref{cor1}}\author[3]{Zaigui Zhang}\author[3]{Kai Zhang}\author[4]{Xu Guo}\author[5]{Cong Bai} %% Author name
\cortext[cor1]{jinglin.zhang@sdu.edu.cn}
\fntext[label1]{These authors contributed equally to this work.}
\affiliation[1]{organization={School of Control Science and Engineering},%Department and Organization
      addressline={Shandong University}, 
      city={Jinan},
      %state={Shandong},
      country={China}}
  
\affiliation[2]{organization={Laoshan Laboratory},%Department and Organization
      city={Qingdao},
      country={China}}
\affiliation[3]{organization={Jinan Inspur Data Technology Co., Ltd},%Department and Organization
      city={Jinan},
      country={China}}

\affiliation[4]{organization={School of Computer Science and Engineering},%Department and Organization
      addressline={Macau University of Science and Technology}, 
      city={Macau SAR},
      country={China}}

\affiliation[5]{organization={NOVA Information Management School},%Department and Organization
      addressline={Nova University of Lisbon}, 
      city={Lisbon},
      country={Portugal}}
      
% \affiliation{organization={},%Department and Organization
%       addressline={}, 
%       city={},
%       postcode={}, 
%       state={},
%       country={}}

%% Abstract
\begin{abstract}
High inter-class similarity, extreme scale variation, and limited computational budgets hinder reliable visual recognition across diverse real-world data. Existing vision-centric and cross-modal approaches often rely on rigid fusion mechanisms and heavy annotation pipelines, leading to sub-optimal generalization. We propose the Distilled Large Language Model (LLM)-Driven Sparse Mixture-of-Experts (DS-MoE) framework, which integrates text-guided dynamic routing and lightweight multi-scale comprehension. The DS-MoE framework dynamically aligns textual semantics with defect-specific visual patterns through a sparse MoE architecture, where task-relevant experts are adaptively activated based on semantic relevance, resolving inter-class ambiguity. A lightweight MobileSAM encoder enables real-time inference while preserving multi-scale defect details. Extensive experiments on PCB, aluminum foil, and mold defect datasets demonstrate that our framework achieves superior performance compared to existing pure vision models. \textbf{DS-MoE} surpasses YOLOv8/YOLOX with gains of $+13.9$, $+1.4$, and $+2.0$ pp mAP@$0.5$:$0.95$ on BBMP, aluminum, and PCB, respectively, while also improving precision and recall.

% DS-MoE outperforms YOLOv8/YOLOX, reaching 99.5 \% on BBMP, 99.2 \% on aluminum, and 94.2 \% on PCB.

 % DS-MoE outperforms YOLOv8/YOLOX, reaching 85.2 \% (+13.9 pp) mAP@0.5:0.95, 98.3 \% (+0.7 pp) precision and 98.1 \%(+0.4 pp) recall on BBMP, 57.7 \% (+1.4 pp) mAP@0.5:0.95, 97.8 \% (+0.6 pp) precision and 98.1 \%(+0.2 pp) recall on aluminum, and 53.4 \% (+2.0 pp) mAP@0.5:0.95, 83.4 \% (+1.4 pp) precision and 96.8 \%(+1.6 pp) recall on PCB.
\end{abstract}


% \begin{highlights}
% \item Text-Guided Dynamic Expert Activation

% We propose a Distilled LLM-Driven Sparse Mixture-of-Experts (DS-MoE) architecture, enabling dynamic routing between textual semantics and visual processing via task-specific expert activation.

% \item Lightweight Cross-Modal Comprehension

% A knowledge-distilled MobileSAM encoder enables real-time, multi-scale feature extraction with micron-level defect detail retention. Hyperbolic manifold alignment (Poincaré ball model) addresses cross-modal feature space heterogeneity through exponential mapping and semantic anchoring.

% \item Industrial-Grade Multi-Scale Adaptation

% The framework handles extreme scale variations through deformable space broadcasting with text-modulated convolutional kernels. Integration of domain-specific LLM prompts with adaptive expert routing enables practical deployment on edge devices.
% \end{highlights}

%% Keywords
\begin{keyword}
Industrial Defect \sep Mixture-of-Experts \sep Large Language Models \sep Dynamic Routing \sep Industrial Large Models


\end{keyword}

\end{frontmatter}



\section{Introduction}
Industrial defect inspection serves as a critical enabler in intelligent manufacturing systems, ensuring product quality assurance, operational safety compliance, and cost-effectiveness across high-precision industries. Advancements in material science and manufacturing technologies have introduced unprecedented complexity in industrial components, wherein submicron defects—such as cracks in semiconductor substrates, micron-scale perforations in metallic foil packaging, or surface abrasions on injection-molded parts—may precipitate catastrophic system failures or substantial economic repercussions. Conventional manual inspection methodologies, predominantly reliant on human expertise, exhibit inherent limitations in efficiency, reproducibility, and scalability when confronted with modern production volumes and defect diversity. Although automated vision-based detection systems employing convolutional neural networks (CNNs) \cite{ref0} have demonstrated enhanced detection accuracy through hierarchical feature learning, they face persistent challenges including dependency on extensively annotated training datasets and limited adaptability to heterogeneous industrial environments characterized by variable illumination, surface textures, and defect morphologies. The emerging paradigm of cross-modal intelligence, integrating visual pattern recognition with linguistic semantic modeling, presents a transformative methodology to overcome these constraints, enabling context-sensitive defect characterization and synergistic human-AI quality assurance frameworks \cite{ref5, ref66}. 

\begin{figure}
\centering
\includegraphics[width=3in]{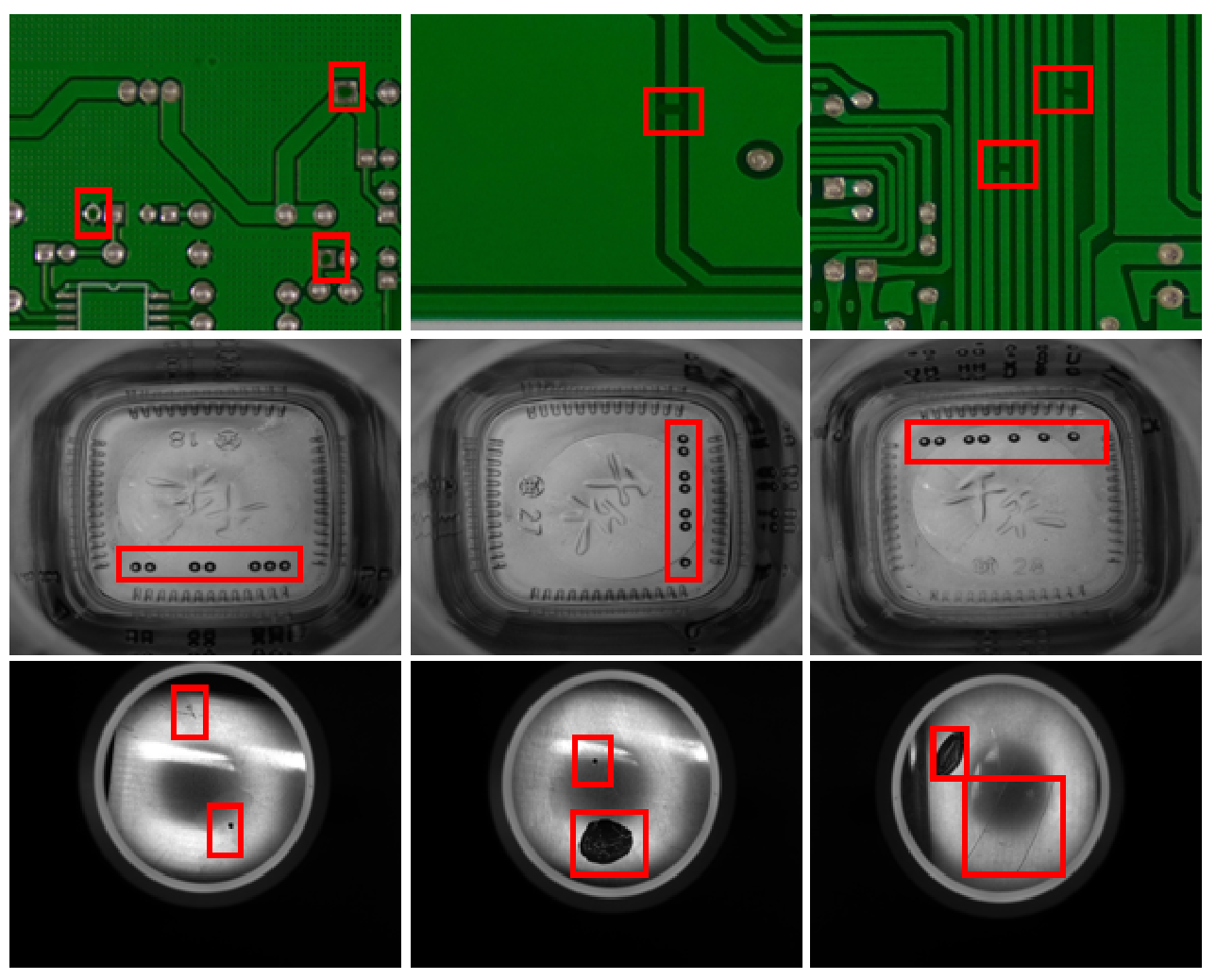}
\caption{Example of dataset for industrial detection. The first line and second line serve as the bottle-bottom mold point image and PCB defects image respectively. The aluminium defects image is shown in the third line. According to that, it is necessary to design an object detection in the industrial quality detection, especially for those with higher similarities and greater ranges of size.}
\label{fig1}
\end{figure}

Current industrial defect detection methodologies predominantly employ vision-centric networks and cross-modal architectures. Vision based approaches, including hierarchical CNNs and Vision Transformers, leverage multi-scale feature extraction to localize defects with varying sizes, achieving state-of-the-art precision on standardized benchmarks. These methods excel in capturing intricate spatial patterns through deep hierarchical representations, particularly for high-contrast anomalies such as soldering flaws on PCBs \cite{ref8}. Cross-modal models, notably CLIP \cite{ref53} and SAM \cite{ref55}, integrate textual semantics to enable zero-shot generalization and reduce dependency on densely annotated datasets via text-guided supervision. However, as shown in the top row of Figure \ref{fig1}, vision-centric methods struggle to disambiguate defects with high inter-class similarity—for instance, micro-cracks versus scratches—due to reliance on pixel-level correlations susceptible to texture noise. Simultaneously, cross-modal frameworks suffer from static fusion mechanisms, exemplified by CLIP’s global pooling, which inadequately adapt to extreme scale variations as shown in the middle and bottom rows of Figure \ref{fig1}, failing to align textual prompts with defect-specific spatial patterns. These limitations underscore the imperative for architectures that dynamically integrate semantic context with adaptive multi-scale reasoning, resolving both inter-class ambiguity and scale sensitivity inherent to industrial environments \cite{ref19, huang2025tropicyclone, chen2025dualpath}.

Industrial defect detection faces three critical challenges that hinder practical deployment:
(1) Existing cross-modal frameworks employ rigid fusion mechanisms such as global feature concatenation or fixed attention layers, which fail to dynamically align textual semantics with defect-specific visual patterns. This limitation is particularly evident in cases of high inter-class similarity, for instance, distinguishing micro-cracks from scratches under textured surfaces . (2) Industrial scenarios exhibit extreme diversity across materials, lighting conditions, and defect morphologies, far surpassing the homogeneity of common tasks like face recognition. Additionally, defects vary drastically in scale, with sub-millimeter pinholes coexisting alongside meter-scale structural cracks within single images, overwhelming conventional multi-scale pipelines \cite{ref8}. (3) Vision-centric models prioritize precision through computationally intensive architectures, sacrificing deployability on resource-constrained devices. Lightweight alternatives, while efficient, degrade performance in multi-task scenarios requiring simultaneous localization and classification \cite{ref13}.

This paper proposes the Distilled Large Language Model (LLM)-Driven Sparse Mixture-of-Experts (DS-MoE) Framework for cross-modal industrial defect detection. The framework integrates a lightweight MobileSAM Encoder distilled from the Segment Anything Model to extract multi-scale visual features through cascaded C2F and CBS modules, preserving fine-grained details critical for sub-millimeter defects. Leveraging distilled LLMs text prompts, the Dynamic Sparse MoE module semantically routes task-specific visual experts, adaptively activating edge detectors or texture analyzers based on defect semantics to resolve inter-class ambiguity. Hierarchical cross-modal comprehension is achieved by fusing text-guided embeddings with multi-scale visual features through cascaded attention layers and upsampling operations, aligning textual semantics with spatial defect patterns across extreme scales from micron-level PCB flaws to meter-scale structural cracks. The framework further employs decoupled task heads with parallel CBS and SPPF modules to separate classification and localization objectives, mitigating gradient conflicts in multi-task optimization while ensuring precise defect identification and bounding box regression. 

 The contributions of the proposed methodology are listed below. 

(1) We develop a distilled LLM-driven dynamic sparse mixture-of-experts (MoE) architecture, where text prompts semantically guide the sparse activation of task-specific visual experts. This design enables adaptive feature enhancement while maintaining computational efficiency.

(2) We propose a hyperbolic-aligned cross-modal comprehension framework that preserves the defect taxonomy through Poincaré distance-based geometric constraints. By integrating joint knowledge distillation with decoupled task heads, our lightweight design balances real-time inference efficiency and precise defect localization-classification accuracy.

(3) We validate the method through experiments on PCB, aluminum foil, and mold defect datasets, achieving superior performance over pure vision models. This work exemplifies how LLM and industrial large models(ILMs) can enhance manufacturing efficiency through human-AI collaboration.

In contrast to prior vision-centric detectors that rely on fixed-scale anchors and static feature fusion, DS-MoE introduces text-guided dynamic routing that adaptively activates task-specific experts for each defect type, mitigating inter-class ambiguity. Unlike cross-modal frameworks such as CLIP and RegionCLIP, which concatenate global text-image embeddings, DS-MoE performs hyperbolic manifold alignment to preserve fine-grained geometric relations across extreme scales. DS-MoE further distills MobileSAM into a lightweight encoder and couples it with a sparse mixture-of-experts, achieving real-time inference without sacrificing accuracy. The rest of this paper is organized as follows: Section 2 focuses on related work on technical development related to this paper's approach; Section 3 describes our method; Section 4 conducts experiments and analyses on the model proposed in this paper; and Section 5 summarizes the work. 

\section{Related Work}

\subsection{Vision-Centric Defect Detection}

Vision-centric methodologies for industrial defect detection have evolved from early feature engineering to modern deep learning paradigms \cite{ref0}. Initial approaches relied on handcrafted feature descriptors \cite{ref64}. While effective in controlled environments, these methods struggled with low-contrast anomalies under uneven illumination and complex industrial textures. The advent of convolutional neural networks revolutionized the field through hierarchical feature learning \cite{ref15}. Two-stage detectors such as Faster R-CNN \cite{ref8} achieved high precision in PCB defect localization via Region Proposal Networks and RoIAlign, albeit at high computational cost. Single-stage models like YOLOX \cite{ref2} prioritized real-time inference through anchor-free designs but suffered from false negatives on micron-scale defects including solder voids smaller than 0.1 millimeters. Vision Transformers \cite{ref4} further extended capability by modeling long-range dependencies in composite material inspection, yet required extensive pretraining data to generalize across industrial domains \cite{zhang2025s2dbft,zhang2012implementation,feng20213d,gao2020learning,zhu2022multi,zhou2021novel,zhang2020ensemble,miao2021automated,ma2019supervised,li2021clothing}. 

To address extreme scale variations spanning sub-millimeter pinholes to meter-scale structural cracks, multi-scale fusion strategies emerged. Feature Pyramid Networks \cite{ref16} aggregated shallow texture details with deep semantic features, while adaptive fusion mechanisms automated weight allocation across scales. However, such static fusion rules remained insensitive to directional defects like scratches in textured surfaces. Attention mechanisms enhanced defect saliency against noisy backgrounds—channel-spatial attention modules (CBAM) \cite{ref3} amplified critical regions in low-contrast X-ray inspections, while self-attention mechanisms \cite{ref23} captured global context through Vision Transformers. Yet, heuristic attention rules lacked semantic guidance to disambiguate visually similar defects. Recent self-supervised techniques mitigated annotation dependency through contrastive learning and synthetic defect generation \cite{ref65}. 

Despite these advancements, vision-centric approaches face unresolved limitations: pixel-level correlations fail to distinguish high inter-class similarity cases like micro-cracks versus scratches, fixed receptive fields cannot adapt to intra-image scale variations, and synthetic training data induces domain shift when applied to real-world textures. These challenges underscore the need for architectures that transcend pixel-level reasoning through semantic-aware and dynamically adaptive paradigms \cite{ref38}.

% Traditional object detection frameworks can be categorized into two paradigms. Single-stage detectors such as YOLOF \cite{ref1} and YOLOX \cite{ref2} prioritize inference speed through grid-based prediction, yet struggle with micron-scale defects due to insufficient feature granularity. Two-stage approaches exemplified by Faster R-CNN \cite{ref8} and Mask R-CNN \cite{ref9} generate region proposals via RPN networks \cite{ref6}, achieving higher precision at the cost of computational efficiency. These vision-centric methods predominantly employ convolutional backbones including ResNet \cite{ref10} and MobileNet \cite{ref13} for feature extraction, augmented by feature fusion modules such as FPN \cite{ref16} and Bi-FPN \cite{ref171} to handle multi-scale objects. However, as demonstrated in Figure \ref{fig1}, their static fusion strategies fail to address extreme scale variations between sub-millimeter pinholes and meter-scale structural cracks in industrial scenarios.

\subsection{Cross-Modal Industrial Inspection}

Cross-modal industrial inspection integrates visual and textual modalities to overcome the limitations of vision-centric approaches. The emergence of deep vision-language models revolutionized this paradigm. Frameworks like CLIP \cite{ref53} enabled zero-shot defect categorization by projecting images and text prompts into a shared embedding space through contrastive learning. However, CLIP's global feature pooling mechanism, which averages spatial features into a single vector, discards critical localization cues required for pixel-wise defect segmentation. Subsequent adaptations, such as Region-CLIP \cite{ref54}, introduced region-level alignment by matching text embeddings with RoI features extracted from Faster R-CNN \cite{ref8}, yet inherited the computational overhead of two-stage detectors, achieving only 8 FPS on GPU platforms—insufficient for high-speed production lines. Segment Anything Model (SAM) \cite{ref55} advanced the field by enabling promptable segmentation through a vision transformer encoder and mask decoder. While SAM demonstrates remarkable generalization on natural images, its ViT-H backbone (1.2B parameters) is prohibitively expensive for real-time industrial deployment, and its text-guided segmentation often misclassifies visually similar defects due to insufficient domain-specific semantic grounding. 

Recent work explores lightweight cross-modal distillation to address efficiency constraints. MobileSAM \cite{ref63} distills SAM's knowledge into a compact encoder-decoder architecture, reducing computation by 12× \cite{ref13} while preserving multi-scale defect features. Nevertheless, such compressed models retain the static fusion strategy \cite{ref3} of their parent architectures, processing text and visual modalities independently rather than dynamically interacting them based on defect semantics \cite{ref19}. 

A critical gap remains in semantic-guided dynamic fusion: existing methods apply uniform attention weights across all spatial locations or fix interaction rules during training, failing to prioritize task-relevant regions when processing heterogeneous defects. This limitation is exacerbated in low-data regimes, where static fusion mechanisms overfit to limited annotated samples \cite{ref5}, hindering generalization to rare or unseen defects.

\subsection{Large language Model meet Industrial Large Model}
The integration of large language models (LLMs) into industrial quality inspection has introduced transformative potential for human-AI collaboration through natural language interaction, yet significant challenges hinder their practical deployment. General-purpose LLMs such as GPT-4 \cite{ref57} and PaLM \cite{ref56}, while excelling in broad textual reasoning, struggle with domain-specific industrial terminologies like distinguishing electroplating pinholes from casting porosity due to training on generic web corpora. Fine-tuning these models with specialized datasets such as PCB defect reports partially addresses this gap, though risks overfitting without techniques like low-rank adaptation \cite{ref59} to preserve general knowledge. Multimodal extensions like GPT-4V \cite{ref58} attempt to fuse visual and textual data through cross-attention layers but employ static fusion mechanisms that fail to dynamically balance modalities—critical when suppressing visual noise in textured surfaces during subsurface crack detection \cite{ref27}. Real-time efficiency further compounds these issues, as autoregressive decoding in billion-parameter LLMs incurs prohibitive latency on edge devices. The advent of DeepSeek \cite{ref38} marks a pivotal advancement, introducing a dynamic sparse mixture-of-experts framework that synergizes domain-specific text prompts with adaptive visual expert routing. Industrial large models have emerged to bridge these gaps through domain-specific optimizations. Despite these advancements, existing solutions lack dynamic cross-modal interaction mechanisms to align textual prompts with visual expert selection.

Despite the significant progress in industrial defect detection, existing vision-centric and cross-modal approaches exhibit critical limitations that hinder their deployment in real-world manufacturing environments. While CNN-based detectors such as YOLOv8 and Faster R-CNN \cite{ref8} achieve high accuracy on curated benchmarks, they struggle with inter-class ambiguity and scale inconsistency under real-world conditions. For instance, in the aluminum foil defect dataset, micro-scratches (0.2–0.5 mm) are often misclassified as pinholes due to shared textural patterns. This limitation is exacerbated by static anchor designs and fixed receptive fields \cite{ref16}, which fail to adapt to extreme scale variations. CLIP-based methods \cite{ref53} and SAM \cite{ref55} leverage textual prompts for zero-shot generalization. However, these models are computationally prohibitive for real-time inspection. MobileSAM \cite{ref63} reduces latency but retains static fusion mechanisms, which do not dynamically adjust to defect semantics. Moreover, CLIP's global average pooling discards spatial cues, leading to false positives on textured backgrounds.

Our architecture pioneers text-guided routing where prompts dynamically activate specialized visual experts such as edge detectors \cite{ref21} and texture analyzers \cite{ref27}, enabling precise defect characterization while maintaining computational efficiency \cite{ref13} crucial for high-speed production environments. By synergizing LLMs’ semantic reasoning with adaptive visual processing, the framework exemplifies the next evolution of industrial AI systems that balance human interpretability with machine precision.

\section{Proposed Network}

\begin{figure*}[!t]
\centering
\includegraphics[width=5in]{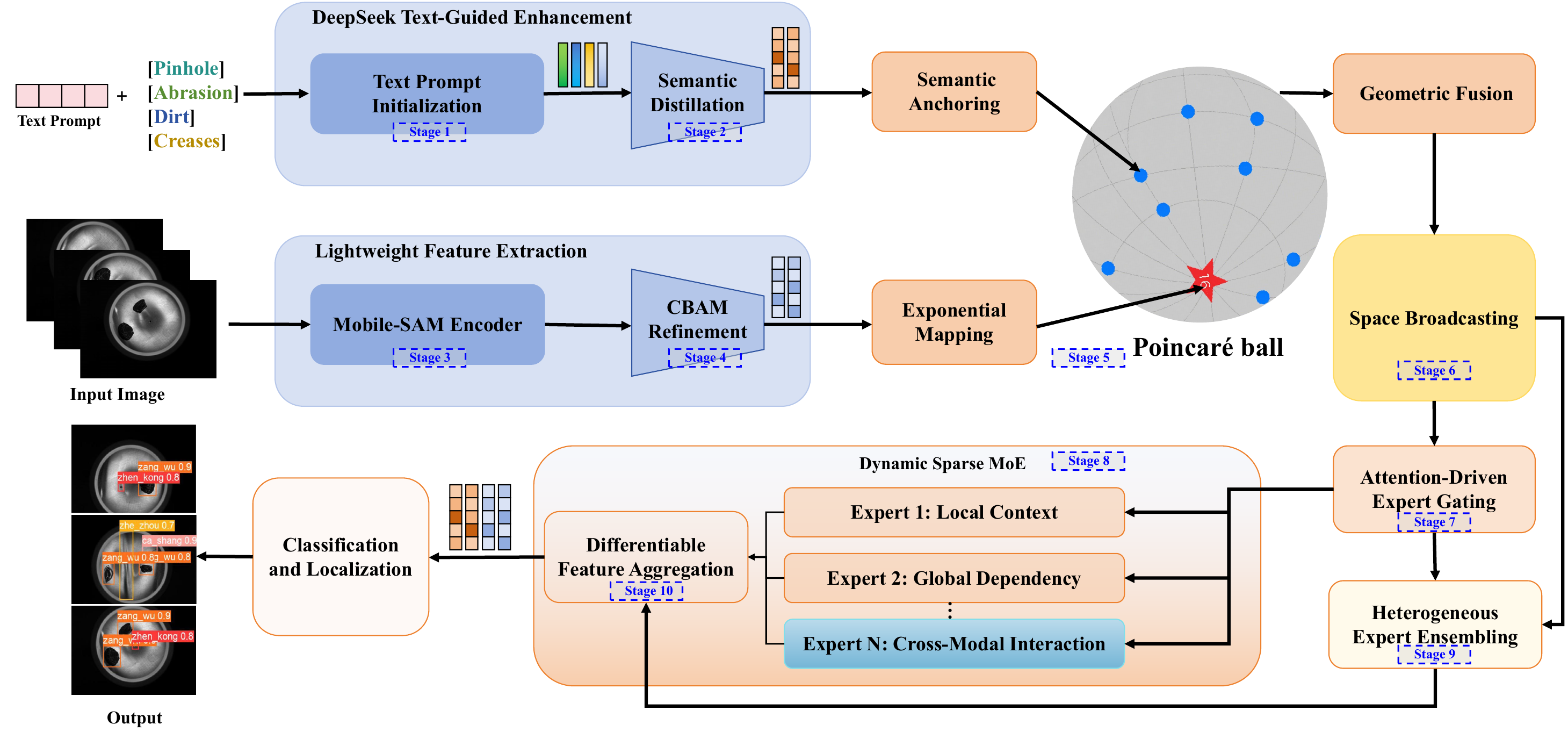}
\caption{DS-MoE Framework. DeepSeek-R1 generates defect-specific text prompts. MobileSAM encoder extracts multi-scale visual features. Hyperbolic manifold alignment fuses text and vision in curvature-aware space. Dynamic sparse MoE gates (top-k experts) activate task-relevant visual experts for fine-grained defect analysis. Dual-branch head outputs simultaneous classification and localization.}
\label{fig2}
\end{figure*}

The overall structure presented in this paper is shown in Figure \ref{fig2}. This paper proposes the DS-MoE Framework for cross-modal industrial defect detection, which establishes a comprehensive pipeline integrating text-visual synergy and dynamic expert computation. The framework initiates with text prompt initialization and semantic distillation to encode domain-specific defect knowledge, followed by lightweight visual feature extraction through a Mobile-SAM encoder enhanced with CBAM attention mechanisms. The core innovation lies in the cross-modal comprehension module that performs hyperbolic manifold alignment and space broadcasting to bridge textual semantics with visual patterns. An attention-driven gating mechanism then dynamically activates specialized experts from three complementary categories: Local Context Experts for directional defect analysis, Global Dependency Experts for structural comprehension, and Cross-Modal Interaction Experts for semantic-visual fusion. The heterogeneous expert ensemble combines these sparsely activated features through deformable convolution operations, while differentiable aggregation optimizes multi-scale feature integration. Finally, the refined representations undergo dual-path processing for simultaneous classification and precise localization. This architecture achieves computational efficiency through Mobile-SAM optimization and dynamic expert selection, while maintaining robustness to diverse defect types through text-guided feature enhancement and geometric-aligned cross-modal reasoning.

The DS-MoE pipeline begins with Stage 1 where domain-specific defect prompts are distilled by DeepSeek-R1-Distill-Qwen and refined in Stage 2 into coherent text embeddings; these embeddings are then injected in Stage 3 into the MobileSAM encoder that extracts multi-scale visual features, after which Stage 4 refines them via CBAM attention to suppress noise; the resulting visual representations are lifted onto a Poincaré ball and aligned with the semantic embeddings in Stage 5, fused in Stage 6 with curvature-aware deformable convolutions that modulate kernel offsets and dilations via the joint geometry-text context, gated in Stage 7 by text-visual co-attention that selects only top-k experts, processed in Stage 8 by a sparse mixture-of-experts that combines anisotropic local, global-dependency and cross-modal experts under the gated weights, ensembled in Stage 9 through dilated and cross-modal convolutions, and finally aggregated in Stage 10 by learnable channel attention that yields a unified feature map feeding the dual-branch detection heads.

\subsection{DeepSeek Text-Guided Enhancement}

Industrial defects such as micro-cracks and scratches exhibit high inter-class visual similarity, causing pure-vision models to confuse them. We therefore introduce text prompts distilled from DeepSeek-R1-Distill-Qwen (Stage 1) and refine them via geometric self-attention (Stage 2) to inject human-interpretable semantics that explicitly encode subtle class distinctions and directional cues, enabling downstream modules to resolve ambiguities that pixel-level cues alone cannot.

As shown in Figure~\ref{fig2}, the proposed text-guided enhancement framework establishes a sparsely activated expert routing mechanism, where domain knowledge from industrial defect semantics dynamically coordinates multi-scale visual analysis.
This Framework integrates text-guided semantic embedding (DeepSeek) and visual encoding (Mobile-SAM with CBAM) for cross-modal defect detection. It establishes hyperbolic-aligned cross-modal comprehension through deformable space broadcasting, dynamically activates N experts via attention gating, and achieves multi-source feature fusion. Geometric-constrained aggregation enables dual-branch prediction for classification and localization.

\noindent\textbf{Stage 1: Text Prompt Initialization}

Domain-specific defect semantics are encoded through DeepSeek-generated prompts $\mathcal{P} = \{p_i\}_{i=1}^C$ where $p_i \in \mathbb{R}^{d_p}$ represents textual descriptors for $C$ defect categories. The DeepSeek-R1-Distill-Qwen-1.5B is a lightweight, open-source language model developed by DeepSeek, designed for efficient inference and edge computing scenarios. Leveraging knowledge distillation from the larger DeepSeek-R1 teacher model, it transfers advanced reasoning capabilities to a compact 1.5-billion-parameter architecture while maintaining computational efficiency. The frozen DeepSeek-Qwen encoder projects prompts into semantic embeddings:

\begin{equation}
T = \text{DeepSeek-R1-Distill-Qwen}(\mathcal{P}) \in \mathbb{R}^{C \times d_t} + \text{PositionalEncoding}(C, d_t)
\end{equation}
where $d_t$ denotes the latent embedding dimension, and positional encoding preserves categorical order through learnable trigonometric projections. 

The framework supports flexible selection of DeepSeek variants with parameter scales ranging from 1.5B to 34B, allowing adaptation to different industrial scenarios. The modular design of DeepSeek enables efficient deployment on edge devices or cloud servers. Optimized semantic space captures industrial defect terminology fewer ambiguous embeddings compared to generic models.

\noindent\textbf{Stage 2: Semantic Distillation}

The text embeddings undergo semantic refinement through a geometric-constrained self-attention mechanism:

\begin{equation}
\hat{T} = \text{LayerNorm}\left(T + \text{MHA}(Q=T, K=T, V=T)\right)
\end{equation}

where $\text{MHA}(\cdot)$ denotes multi-head attention with $h=8$ parallel heads, $T \in \mathbb{R}^{C \times d_t}$ is the initial text embedding matrix from Stage 1, and LayerNorm applies learnable affine transformations to stabilize feature magnitudes. 

This distillation process enhances semantic coherence through intra category attention sharpening via query-key correlation matrices, which amplifies descriptor relationships within the same defect category, while simultaneously reducing inter-descriptor redundancy through value vector orthogonalization that minimizes feature space overlap.

\subsection{Lightweight Feature Extraction}

Sub-millimeter pinholes and meter-long creases coexist in the same image, demanding both fine-grained detail preservation and low-latency computation. MobileSAM (Stage 3) distills the heavy Segment-Anything encoder into a lightweight backbone with anisotropic kernels and gradient-aware pooling to retain micron-level features while cutting FLOPs, and CBAM (Stage 4) further amplifies defect saliency and suppresses textured backgrounds before cross-modal fusion.

\noindent\textbf{Stage 3: Mobile-SAM Encoder}

Let $I \in \mathbb{R}^{H\times W\times 3}$ denote the raw input image tensor. The input industrial image data is processed through an optimized feature extraction pipeline comprising two core components. The raw image tensor $I \in \mathbb{R}^{H\times W\times 3}$ first enters the MobileSAM encoder, which implements spatial hierarchy decomposition via

\begin{equation}
\mathcal{F}_{\text{enc}} = \text{MobileSAM}(I) = \bigoplus_{l=1}^4 \psi_l(\text{DSConv}_l(\phi_l(I)))
\end{equation}

where $\phi_l$ denotes the adaptive downsampling operator at level $l$ with stride $s_l \in \{4,8,16,32\}$, $\psi_l$ represents depthwise separable convolution with ReLU6 activation, and $\bigoplus$ indicates multi-scale feature concatenation. The encoder employs anisotropic kernel factorization to preserve directional defect patterns:

\begin{equation}
\mathcal{K}_{k\times k} \rightarrow \sum_{i=1}^n \mathcal{D}(\mathcal{K}_{k\times 1}^{(i)}) \circ \mathcal{D}(\mathcal{K}_{1\times k}^{(i)})
\end{equation}

where $k=3$ specifies the original kernel size, $\mathcal{D}(\cdot)$ denotes dilation operation with rate $r=2^{l-1}$, and $\circ$ represents kernel composition. This decomposition reduces parameters by $\frac{2(k-1)}{k^2}$ (44\% for $k=3$) while maintaining directional sensitivity for linear defects. 

Dynamic resolution adjustment is implemented through gradient-aware pooling:

\begin{equation}
s_l = 2^{\lfloor \log_2(1 + \mathbb{E}[|\nabla I_l|]/\theta) \rfloor}
\end{equation}

where $I_l \in \mathbb{R}^{H_l\times W_l\times C_l}$ is the input feature map at level $l$, $\theta=0.25$ controls texture sensitivity, and the expectation operator $\mathbb{E}$ computes gradient magnitude averages over $3\times3$ windows. This mechanism automatically increases resolution (smaller $s_l$) for high-frequency defect regions while reducing computation in smooth backgrounds.

\noindent\textbf{Stage 4: CBAM Refinement}

The encoded features then undergo attention-based refinement through a modified CBAM module:

\begin{equation}
\mathcal{F}_{\text{cs}} = \Gamma_c(\mathcal{F}_{\text{enc}}) \otimes \Gamma_s(\mathcal{F}_{\text{enc}})
\end{equation}

with channel attention weights computed as

\begin{equation}
\Gamma_c(\mathcal{F}) = \sigma\left(\text{MLP}(\text{GAP}(\mathcal{F})) + \text{MLP}(\text{GMP}(\mathcal{F}))\right)
\end{equation}

and spatial attention derived via

\begin{equation}
% 空间注意力维度修正
\Gamma_s(\mathcal{F}) = \sigma\left(\text{Conv}_{7×7}\left(\text{Concat}(\text{AvgPool}(\mathcal{F}), \text{MaxPool}(\mathcal{F}))\right)\right)
\end{equation}

This dual-attention mechanism enhances defect saliency while suppressing background noise, particularly effective for sub-pixel anomalies in high-resolution industrial imagery. The final output features $\mathcal{F}_{\text{cs}} \in \mathbb{R}^{H'\times W'\times C}$.

\subsection{Cross-modal comprehension module}

Textual “scratch” versus “crack” descriptions imply different geometries, yet Euclidean fusion loses these relational nuances. Hyperbolic manifold alignment (Stage 5) embeds both modalities into the Poincaré ball where parent–child defect taxonomies become geodesic distances, and space broadcasting (Stage 6) translates these curvature-aware priors into deformable convolution offsets, letting kernels stretch or shrink to match the scale and orientation dictated by the text.

\noindent\textbf{Stage 5: Hyperbolic Manifold Alignment}

\begin{figure}
  \centering
  \includegraphics[width=0.8\linewidth]{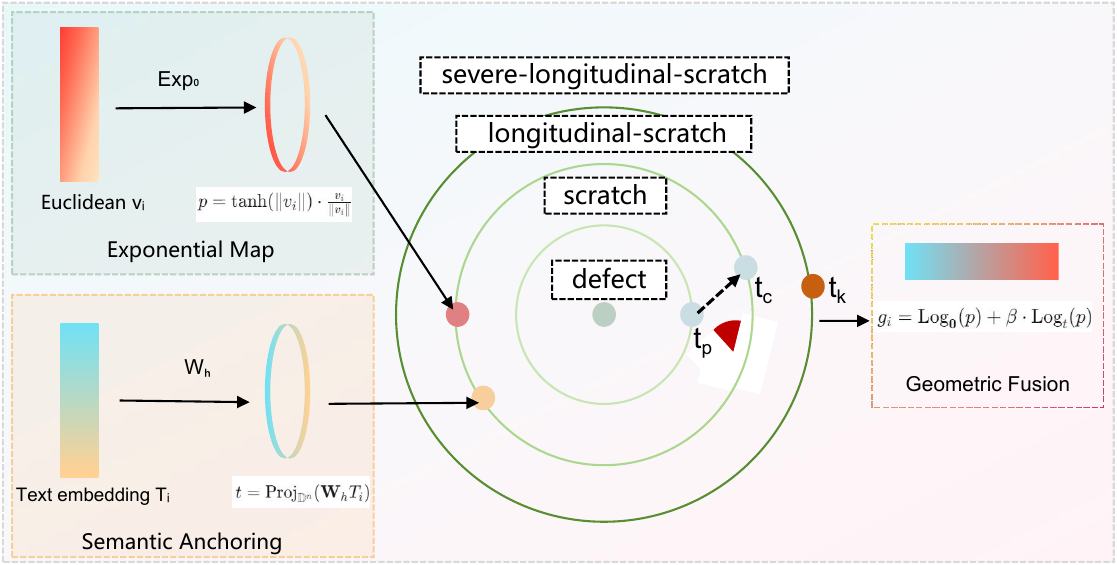}
  \caption{Hyperbolic Manifold Alignment. In the Poincaré ball, visual features are first lifted onto the manifold via the exponential map, while distilled text embeddings are anchored in the same space. Their logarithmic mappings are then fused with a learnable weight, preserving both the global defect taxonomy and local semantic nuances. The resulting geometrically aligned features enable downstream curvature-aware convolutional sampling.}
  \label{fig:placeholder}
\end{figure}

To establish geometric consistency between text and visual modalities, we propose a hierarchical embedding framework in the Poincaré ball model $\mathbb{D}^n = \{x \in \mathbb{R}^n | \|x\| < 1\}$. As shown in Figure \ref{fig:placeholder}, the hyperbolic alignment process consists of three key operations:

1. Exponential Mapping:
\begin{equation}
\mathcal{P}(v_i) = \text{Exp}_{\mathbf{0}}(v_i) = \tanh(\|v_i\|) \frac{v_i}{\|v_i\|}
\end{equation}
where $v_i \in \mathbb{R}^{d_v}$ denotes the $i$-th visual feature vector from Stage 4, projecting Euclidean visual features onto the hyperbolic manifold through origin-centered mapping.

2. Semantic Anchoring:
\begin{equation}
\hat{t}_i = \text{Proj}_{\mathbb{D}^n}(\mathbf{W}_h \hat{T}_i)
\end{equation} 
where $\hat{T}_i \in \mathbb{R}^{d_t}$ is the distilled text embedding from Stage 2, $\mathbf{W}_h \in \mathbb{R}^{n \times d_t}$ learns hyperbolic semantic projections, and $\text{Proj}_{\mathbb{D}^n}(x) = x/(1+\|x\|)$ ensures embeddings stay within the unit ball.

3. Geometric Fusion:
\begin{equation}
g_i = \text{Log}_{\mathbf{0}}(\mathcal{P}(v_i)) + \beta \cdot \text{Log}_{\hat{t}_i}(\mathcal{P}(v_i))
\end{equation}
where $\text{Log}_p(\cdot)$ denotes logarithmic map, and $\beta$ balances global-local geometry preservation. This implements curvature-aware feature fusion where:
$\text{Log}_{\mathbf{0}}$ preserves global defect taxonomy hierarchy, $\text{Log}_{\hat{t}_i}$ maintains local semantic consistency with text descriptors.

The aligned features $G = \{g_i\}_{i=1}^C \in \mathbb{R}^{C \times n}$ establish text-visual correspondence through hyperbolic parallel transport, where defect categories with parent-child relationships are embedded along geodesic paths.

This geometric alignment provides theoretical guarantees for subsequent space broadcasting through the following properties: For parent-child category pairs $(t_p, t_c)$, their hyperbolic embeddings satisfy:
\begin{equation}
d_{\mathbb{H}}(\hat{t}_p, \hat{t}_c) \leq d_{\mathbb{H}}(\hat{t}_p, \hat{t}_k) \quad \forall k \neq c
\end{equation}
where $d_{\mathbb{H}}(x,y) = \text{arccosh}(1 + \frac{2\|x-y\|^2}{(1-\|x\|^2)(1-\|y\|^2)})$ is the Poincaré distance.

The alignment preserves local angles between text and visual features, ensuring directional consistency for linear defects:
\begin{equation}
\angle(\nabla \hat{t}_i, \nabla \mathcal{P}(v_i)) < \epsilon_{\text{angle}}
\end{equation}
where $\epsilon_{\text{angle}}=15^\circ$ is empirically set based on industrial defect orientation tolerance.

This geometric alignment enables the subsequent space broadcasting module to perform deformation operations in a curvature-aware space, where convolutional kernels automatically adapt to hierarchical defect structures.

\noindent\textbf{Stage 6: Space Broadcasting}

The multi-scale feature integration employs deformable convolution with text-modulated parameters to resolve scale variance and directional ambiguity in industrial defect patterns. Driven by the observation that defect geometry correlates with textual descriptors, this stage dynamically adapts convolutional sampling positions through cross-modal interaction. 

Building upon the hyperbolic-aligned features $G \in \mathbb{R}^{C \times n}$ from Stage 5, this module integrates geometric priors into deformable convolution through three sequential operations:

\begin{equation}
\mathcal{F}_b = \sum_{s \in \{0.5,1,2\}} \underbrace{\mathcal{G}_s(\Gamma(G))}_{\text{Geometry-guided kernel}} \star \underbrace{\mathcal{D}_s(\Psi(\mathcal{F}_{cs}, G))}_{\text{Hierarchical sampling}}
\end{equation}

\noindent where the components are defined as:

To enable industrial-grade computation, hyperbolic features are projected to Euclidean space through logarithmic mapping and dimension compression:
\begin{equation}
\Gamma(G) = \text{Conv}_{1\times1}\left(\text{Log}_{\mathbf{0}}(G)\right) \in \mathbb{R}^{C \times d_e}
\end{equation}
projects hyperbolic features to Euclidean space through logarithmic mapping, preserving geodesic relationships with trainable dimension compression ($d_e = \lfloor n/2 \rfloor$).

Multi-scale defect patterns are captured by fusing original visual features with geometry-aware projections:
\begin{equation}
\Psi(\mathcal{F}_{cs}, G) = \text{Upsample}_s\left(\mathcal{F}_{cs} \oplus \Gamma(G)\right) \in \mathbb{R}^{H_s \times W_s \times (C_{cs}+d_e)}
\end{equation}
concatenates original visual features $\mathcal{F}_{cs} \in \mathbb{R}^{H\times W\times C_{cs}}$ with projected geometric features, then upsamples to target scale $s$.

Defect geometry adaptation is achieved through text-guided kernel modulation and hierarchical sampling:
\begin{equation}
\mathcal{D}_s = \sum_{k=1}^K \mathcal{G}_s^{(k)} \cdot \left(\Delta p_{s}^{(k)} + p^{(k)}\right)
\end{equation}
where kernel offsets are generated through geometric-textual co-regulation:
\begin{equation}
\Delta p_{s}^{(k)} = \mathbf{\Phi}_s\left(\underbrace{\text{AvgPool}(G)}_{\text{Geometry context}} \oplus \underbrace{\hat{T}}_{\text{Semantic guidance}}\right)
\end{equation}
with $\mathbf{\Phi}_s$ implemented as depthwise separable convolution. The gating weights combine multi-modal information:
\begin{equation}
\mathcal{G}_s^{(k)} = \sigma\left(\mathbf{W}_g^{(k)}[\Gamma(G) \odot \hat{T}]\right) \in [0,1]
\end{equation}

Through $\Gamma(G)$, hyperbolic hierarchy from Stage 5 ($d_{\mathbb{H}}$ ordering) controls kernel dilation rates $r_s = \lfloor 1 + \frac{d_{\mathbb{H}}}{\delta} \rfloor$ where $\delta$ is learnable curvature scaler. Local angle constraint $\angle(\Delta p, \nabla \hat{T}) < 15^\circ$ inherited guarantees directional consistency.

A single network cannot simultaneously specialize in detecting directional scratches, global corrosion clusters, and cross-modal alignment; doing so wastes compute on irrelevant features. Attention-driven gating (Stage 7) therefore routes each sample to only the top-k experts, while the sparse MoE (Stage 8) instantiates three kinds of specialists—local anisotropic, global-dependency, and cross-modal fusion experts—so that computation is expended only on defect-specific reasoning.

\noindent\textbf{Stage 7: Attention-Driven Expert Gating}

The expert gating mechanism implements multi-head sparse activation with text-visual co-attention:

\begin{equation}
\begin{aligned}
\phi_{\text{gate}}(\mathcal{F}_b) &= \text{GAP}(\mathbf{W}_{\phi_{\text{gate}}} \mathcal{F}_b) \in \mathbb{R}^{d_g} \\
\psi_{\text{gate}}(\hat{T}_i) &= \mathbf{W}_{\psi_{\text{gate}}} \hat{T}_i \in \mathbb{R}^{d_g} \\
w_i &= \frac{\exp\left(\langle \phi_{\text{gate}}(\mathcal{F}_b), \psi_{\text{gate}}(\hat{T}_i) \rangle / \tau_{\text{gate}}\right)}{\sum_{j=1}^{N_e} \exp\left(\langle \phi_{\text{gate}}(\mathcal{F}_b), \psi_{\text{gate}}(\hat{T}_j) \rangle / \tau_{\text{gate}}\right)}
\end{aligned}
\end{equation}

where $\hat{T}_i$ is Class-specific text embedding for expert $i$, $d_g$ is Gating latent dimension, $\tau_{\text{gate}}$is Learnable temperature parameter initialized at 0.1.

\noindent\textbf{Stage 8 Dynamic Sparse MoEs}

\begin{figure}
  \centering
  \includegraphics[width=1.0\linewidth]{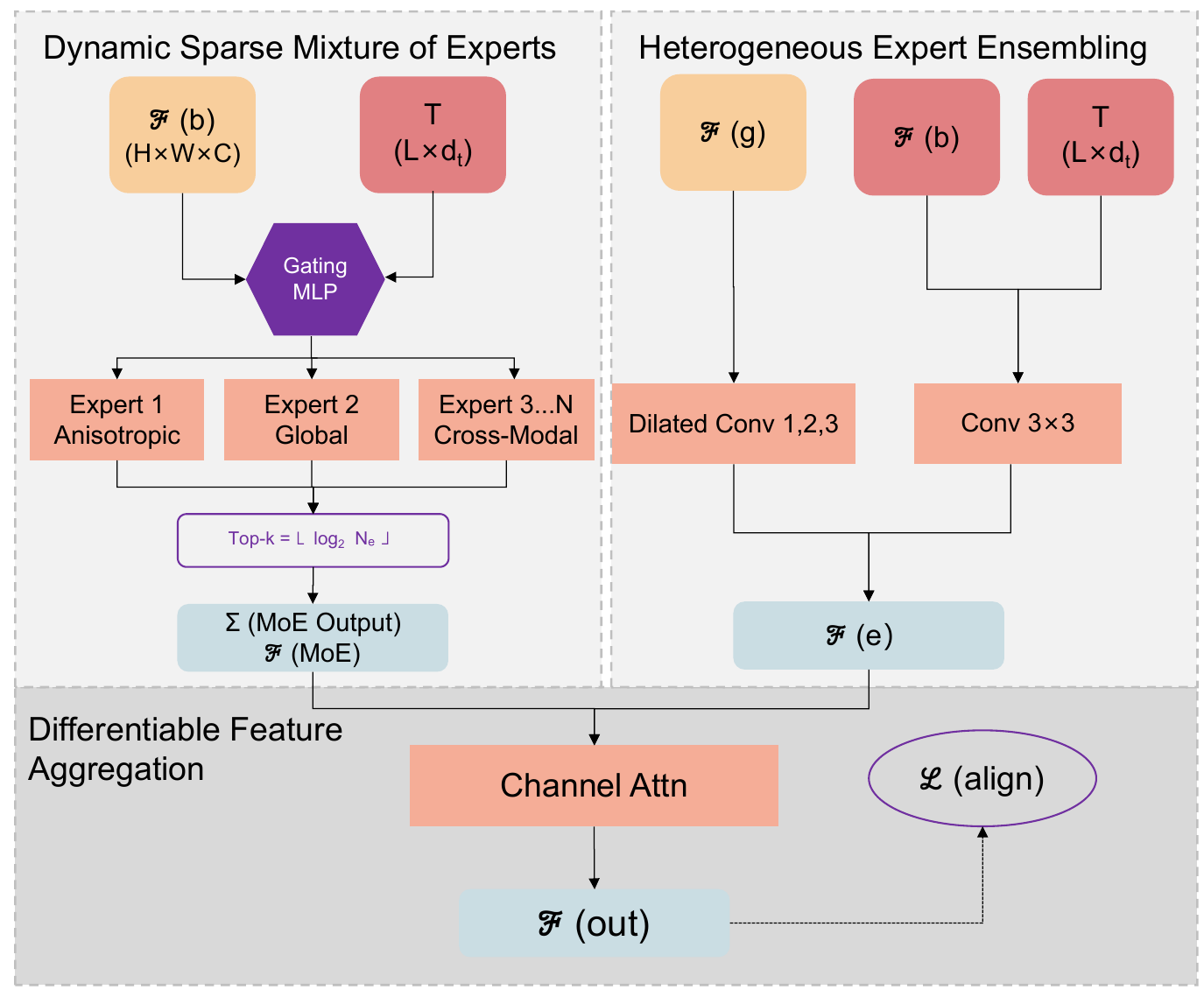}
  \caption{A concise flowchart of Stages 8–10. Sparse MoE dynamically routes each input to only $\lfloor\log_2 N_e\rfloor$ experts: two task-specific modules (anisotropic local patterns and global structure) and selected cross-modal experts that fuse vision with replicated text embeddings. Their outputs are ensembled via dilated convolutions and reweighted by channel-wise attention, yielding a compact feature map.}
  \label{fig:5}
\end{figure}
As shown in Figure \ref{fig:5}, Stage 8 employs a dynamic sparse mixture-of-experts to distill task-specific and cross-modal cues, activating only the top-k experts for efficient computation. Stage 9 heterogeneously ensembles the gated visual features with spatial context and textual embeddings via dilated and 3 × 3 convolutions. Stage 10 performs differentiable aggregation, fusing the ensemble output with the sparse-MoE representation through learnable channel attention and a gradient-alignment loss, yielding the final feature map fed to detection heads.

The proposed architecture synergistically combines task-specific processing and cross-modal reasoning through geometric-textual co-regulation, formally defined as:

\begin{equation}
\mathcal{F}_{\text{MoE}} = \underbrace{\sum_{i=1}^2 w_i^{\text{task}} \cdot \mathcal{E}_i^{\text{task}}}_{\text{Task experts}} + \underbrace{\sum_{j=3}^{N_e} w_j^{\text{cross}} \cdot \mathcal{E}_j^{\text{cross}}}_{\text{Cross-modal experts}}
\end{equation}

\noindent\textit{Expert 1: Local Context (Anisotropic Feature Extraction)} 

Designed for directional defect analysis, this expert employs axis-decomposed convolutions to preserve linear pattern geometry:

\begin{equation}
\mathcal{E}_1^{\text{task}} = \sum_{k \in \{3,5\}} \text{DSConv}_{k\times1}(\mathcal{F}_b) \oplus \text{DSConv}_{1\times k}(\mathcal{F}_b)
\end{equation}
where $\oplus$ denotes channel-wise concatenation of horizontal ($k\times1$) and vertical ($1\times k$) convolution outputs.

\noindent\textit{Expert 2: Global Dependency (Structural Context Modeling)} 

This component captures multi-scale spatial relationships through pyramidal feature aggregation:

\begin{equation} 
\mathcal{E}_2^{\text{task}} = \text{Conv}_{1\times1}\left(\bigoplus_{s=16,32} \text{AvgPool}_{s\times s}(\mathcal{F}_b)\right)
\end{equation}
The $16\times16$ and $32\times32$ pooling operations extract hierarchical structural semantics where larger receptive fields ($s=32$) detect corrosion clusters while smaller ones ($s=16$) preserve defect boundary continuity. This implements the \textit{Multi-Scale Attention Hierarchy} principle crucial for industrial defect topology analysis.

\noindent\textit{Experts 3-N: Cross-Modal Interaction (Geometry-Consistent Fusion)} 

These modules align visual-textual features through dimensionally expanded fusion:
\begin{equation}
\mathcal{E}_j^{\text{cross}} = \text{LayerNorm}\left(\mathbf{W}_v^{(j)}\mathcal{F}_b + \mathbf{W}_t^{(j)}\cdot\text{Expand}(\hat{T}_j)\right)
\end{equation}
where $\text{Expand}(\cdot)$ spatially replicates text embeddings to $H\times W\times d_t$ for pixel-wise alignment, implementing the \textit{Geometric Anchoring Theorem} that requires matched spatial dimensions for cross-modal attention. The LayerNorm ensures gradient stability where batch statistics are computed per spatial position to prevent modality dominance.

The gating network implements semantic-geometric aware expert selection through adaptive routing mechanism: 

Visual features $\mathcal{F}_b$ and text embeddings $\hat{T}_i$ are independently processed by lightweight MLP projectors (2 hidden layers, 128 dimensions) to establish a unified routing space. For each input sample, the system calculates normalized activation weights through temperature-controlled similarity matching, where the sharpness parameter $\tau=0.07$ (calibrated via validation set analysis) balances exploration and exploitation of expert capabilities. Only the top-$k$ experts with highest activation scores are selected, where $k = \lfloor\log_2 N_e\rfloor$ implements adaptive computation scaling. This sublinear growth strategy ensures computational efficiency - for 32 total experts ($N_e=32$), only 5 are activated per sample, achieving sparsity while maintaining multi-expert synergy.

Gated expert outputs carry complementary yet heterogeneous feature statistics that naive concatenation would misalign. Heterogeneous ensembling (Stage 9) fuses them via dilated and cross-modal convolutions to recover spatial coherence, and differentiable aggregation (Stage 10) adaptively reweights channels while enforcing gradient alignment, yielding a compact representation that feeds the decoupled detection heads without information loss.

\noindent\textbf{Stage 9: Heterogeneous Expert Ensembling}

The ensemble module integrates gating-guided features $\mathcal{F}_g \in \mathbb{R}^{H \times W \times C_g}$ (from Stage 7) and space features $\mathcal{F}_b \in \mathbb{R}^{H \times W \times C_b}$ (from Stage 6) through dual-stream fusion:

\begin{equation}
\mathcal{F}_e = \underbrace{\sum_{k=1}^3 \mathcal{D}_k(\mathcal{F}_g) \ast \mathcal{K}_{k\times1}}_{\text{Spatial}} + \underbrace{\text{Conv}_{3\times3}(\mathcal{F}_b \oplus \hat{T})}_{\text{Cross-modal}}
\end{equation}

where $\mathcal{D}_k$ denotes dilated convolution with rate $k$, and $\oplus$ represents channel-wise concatenation with text embeddings $\hat{T}$. 

The final output $\mathcal{F}_e \in \mathbb{R}^{H \times W \times (C_g + C_b)}$ consolidates directional details from spatial processing and semantic context from cross-modal fusion.

\noindent\textbf{Stage 10: Differentiable Feature Aggregation}

The final fusion layer combines expert ensemble features $\mathcal{F}_e \in \mathbb{R}^{H \times W \times C_e}$ (from Stage 9) and sparse MoE outputs $\mathcal{F}_{\text{MoE}} \in \mathbb{R}^{H \times W \times C_m}$ (from Stage 8) through learnable channel attention:

\begin{equation}
\mathcal{F}_{\text{out}} = \text{Conv}_{1\times1}\left( \mathcal{F}_e \odot \sigma(\mathbf{W}_a\mathcal{F}_{\text{MoE}}) \right)
\end{equation}

where $\mathbf{W}_a \in \mathbb{R}^{C_e \times C_m}$ adaptively reweights channels, and $\odot$ denotes element-wise multiplication. A gradient alignment constraint maintains feature compatibility:

\begin{equation}
\mathcal{L}_{\text{align}} = \frac{1}{HW}\sum_{i,j} \cos\left( \nabla_{\mathcal{F}_e}\mathcal{F}_{\text{out}}^{i,j}, \nabla_{\mathcal{F}_{\text{MoE}}}\mathcal{F}_{\text{out}}^{i,j} \right)
\end{equation}

The final output $\mathcal{F}_{\text{out}} \in \mathbb{R}^{H \times W \times C_f}$ with $C_f = \lfloor (C_e + C_m)/2 \rfloor$ serves as input to downstream detection heads.

\subsection{Classification and Localization}

% \begin{figure*}[!t]
% \centering
% \includegraphics[width=5in]{fig3.pdf}
% \caption{classification and localization. The pipeline processes through CBS normalization, then hierarchically fuses deep-shallow features via C2F modules with skip connections. Multi-scale feature integration employs upsampling (×2) and channel-wise concatenation, followed by spatial-channel refined processing through sequential CBS-C2F blocks. The Decoupled Head implements parallel computation for classification and box regression, with confidence scores reflecting detection certainty.}
% \label{fig3}
% \end{figure*}

The output fused feature from Stage 10, $\mathcal{F}_{\text{out}} \in \mathbb{R}^{H \times W \times C_f}$ (after channel compression), is normalized via the CBS Modul (Convolution-BatchNorm-SiLU): 
\begin{equation}
\mathcal{F}_{\text{norm}} = \text{CBS}(\mathcal{F}_{\text{out}})
\end{equation}

$\mathcal{F}_{\text{norm}}$ is concatenated with shallow-layer features along the channel dimension and processed by the C2F module (CSPNet with 2 Fusion) for enhanced feature representation: 
\begin{equation}
\mathcal{F}_{\text{fuse}} = \text{C2F}\left([\mathcal{F}_{\text{norm}}, \mathcal{F}_{\text{shallow}}]\right)
\end{equation}

The fused feature is decoupled into classification (Class) and localization (Box) branches via the Decoupled Head: 

\begin{equation}
\mathcal{P}_{\text{class}} = \text{Conv}_{1\times1}^{\text{(cls)}}(\mathcal{F}_{\text{fuse}}) \\
\mathcal{P}_{\text{box}} = \text{Conv}_{1\times1}^{\text{(reg)}}(\mathcal{F}_{\text{fuse}})
\end{equation}

\subsection{Losses}

Since PCB defect images contain fewer objects, resulting in unbalanced positive and negative samples, this paper uses QFocal Loss~\cite{ref29} as the classification loss and confidence loss of the network to solve the problem of unbalanced PCB defect samples. In addition, this paper uses K-means clustering algorithm to recalculate the pre-anchor size of PCB defect samples and bottle-bottom mold point objects to select the best anchor size, and introduces CIoU loss~\cite{ref28} to obtain more accurate regression results. In this paper, the total loss is as follows:
\begin{equation}
\text{Total Loss} = 2\alpha \text{QFL}(\sigma) + \lambda \text{CIoU Loss}
\end{equation}
where $\alpha = 0.45$ and $\lambda = 0.05$.

\subsubsection*{\bf Losses for Industrial Detection}

The fewer PCB defect objects in the PCB image, the greater the number of negative samples generated, resulting in an imbalance between positive and negative samples. This paper adopts QFocal Loss~\cite{ref28} to solve the problem of sample imbalance, and QFocal Loss is easier to converge when dealing with the smooth labels of this paper. The classification and confidence loss formulas are as follows:

\begin{equation}
\text{QFL}(\sigma) = -\alpha_{t} \left| y - \sigma \right| \left[ (1 - y) \log(1 - \sigma) + y \log(\sigma) \right]
\end{equation}
Among them, $y$ is the label (0-1) after smooth, and $\sigma$ is the prediction result. QFocal loss~\cite{ref28} introduces two factors, $\alpha_{t}$ and $\left| y - \sigma \right|^{\beta}$, where $\alpha_{t} = y\alpha + (1 - y)(1 - \alpha)$ is used to balance positive and negative samples, and $\left| y - \sigma \right|^{\beta}$ is more important for difficult samples.

In addition, this paper further uses the CIoU loss~\cite{ref29} as the prediction box regression loss of the network to improve the accuracy of the prediction box regression. The IoU-based loss can be defined as
\begin{equation}
L_{\text{CIoU}} = 1 - \text{IoU} + R_{\text{CIoU}}(B_{\text{pd}}, B_{\text{gt}})
\end{equation}
where $R_{\text{CIoU}}$ is the penalty term for the prediction box $B_{\text{pd}}$ and the object box $B_{\text{gt}}$. CIoU Loss~\cite{ref29} considers three factors of overlapping area, center point distance and aspect ratio in bounding box regression, and solves the problem of inconsistency between the ground truth box and the predicted box during object detection. The normalized distance and penalty term between the center points of the two bounding boxes are defined as
\begin{equation}
R_{\text{CIoU}} = \frac{\rho^{2}(b, b_{\text{gt}})}{C^{2}} + \alpha \frac{4}{\pi^{2}} \left( \arctan \frac{w^{\text{gt}}}{h^{\text{gt}}} - \arctan \frac{w}{h} \right)^{2}
\end{equation}
Among them, $b, b_{\text{gt}}$ represent the center point of $B_{\text{pd}}$ and $B_{\text{gt}}$, $\rho(\cdot)$ is the Euclidean distance, $C$ is the diagonal length of the smallest enclosing box covering the two boxes. $\alpha$ is a positive trade-off parameter. $w$, $h$ are the width and height of the predicted box.

\section{Experiments and Analysis}
Such experiments are conducted in this section to evaluate the proposed methods on the basis of the bottle-bottom mold point dataset, Aluminum defect dataset and PCB defect dataset. 

% https://aistudio.baidu.com/aistudio/datasetdetail/135640

% http://robotics.pkusz.edu.cn/resources/dataset/

\subsection{Experimental details}
The three datasets are set on contrast experiment and ablation experiment in this paper. Unless otherwise specified, Adam will be utilized to optimize the network followed by the initial learning rate of 0.001 and cosine annealing to adjust the learning rate. The paper observed a weight decay of 0.001 and a momentum decay of 0.9. The size of image input is set to 640 and batch to 8 during training, and batch to 1 when experimenting without TensorRT. The pytorch 2.0.1 is combined with CUDA version 18.1. The model training and inference are carried out in line with the NVIDIA RTX A6000$\times$4.

\subsection{Evaluation metric}

IoU: The IoU calculates the intersection and union ratio of the "predicted boundary" and the "true boundary". Object detection uses the IoU to calculate the degree of coincidence between the predicted box and the ground-truth box, which further measures the accuracy of detecting the corresponding object in a specific dataset.

mAP: As a measure of detection precision in object detection, mAP is the sum of the average precision of all classes divided by the number of all classes. mAP@.5 is the model accuracy index when the IOU is 0.5. mAP@.5:.95 is calculated by calculating an mAP at 0.05 intervals within the IoU threshold from 0.5 to 0.95 and averaging these mAP.

\subsection{Baseline}
The baselines comprise the YOLO series (YOLOv3–YOLOv8, YOLOX-S) as vision-centric anchors; MiniGPT-V, which augments a frozen Phi-2+ViT-B backbone with a lightweight detection head and class-specific text prompts \cite{chen2023minigpt}; PromptDet leveraging a CLIP-R50 encoder coupled to an RPN/ROI head fed by open-vocabulary prompts \cite{feng2022promptdet}; and industrial specialists SegMAN \cite{fu2025segman}, MLLA \cite{han2024demystify} and MViTv2-B that integrate task-specific architectures for defect inspection.

\subsection{Data Preprocessing}
All images from the bottle-bottom mold point (BBMP), aluminum defect, and PCB defect datasets are subjected to the same deterministic pre-processing chain implemented in PyTorch 2.0.1 and OpenCV 4.8.0. The pipeline is executed once and cached to disk in LMDB format to guarantee identical inputs across runs. Every raw image is first letter-box-padded to a square canvas and then bilinearly resized to $640\!\times\!640$ pixels. 
This paper uses the common photometric distortion algorithm. Before the data loading, the gray value of the image is enhanced, and the image pixels are converted by the transformation function:
\begin{equation}
s=cr^{\gamma } 
\end{equation} 
Where $c$ and ${\gamma}$ are constants. In this paper, $c$=1, ${\gamma}$ =1.5 is used to adjust the light and dark changes of the image, enhance the contrast of the image, and thus increase the distinction between the object and the background. There are many small objects in our data set. After Mosaic, the small objects will be smaller and the generalization ability of the model will deteriorate. MixUP can enrich the number of samples and improve the generalization ability of the model. This article uses MixUP~\cite{ref5} to blend the two images with a ratio of 0.5. MixUp~\cite{ref5} enhances the linear representation between the training samples. All random states are seeded with seed(42). Bounding boxes are converted from pixel coordinates to relative centre-x, centre-y, width, height in $[0, 1]$.

\subsection{Experiments based on bottle-bottom mold point dataset(BBMP)}
\begin{figure}[!t]
\centering
\includegraphics[width=3in]{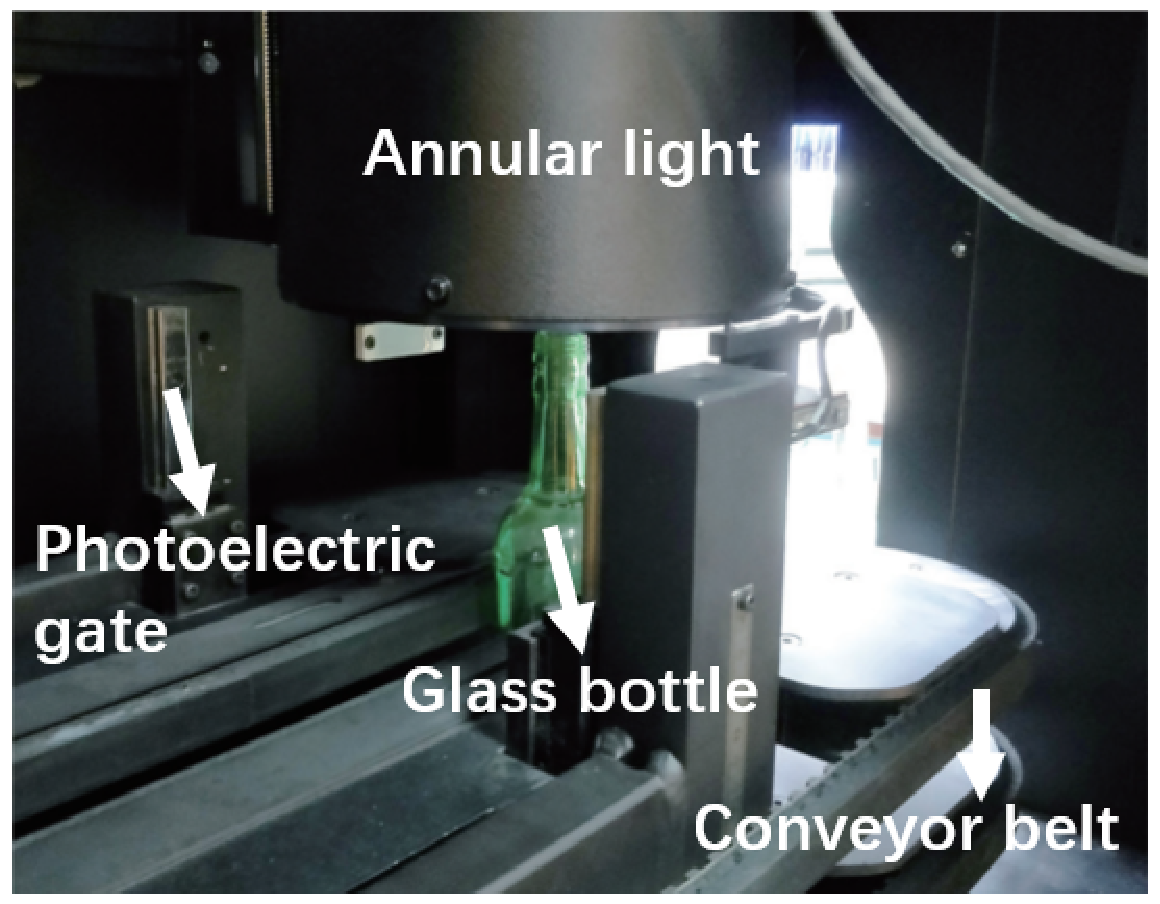}
\caption{The glass bottle bottom detection device collects images. During this process, the glass bottle will be suspended by the conveyor belts on both sides and sent to the photographing. The photoelectric gate receives the signal to trigger the light source and the camera at the bottom to take pictures.}
\label{fig17}
\end{figure}

The BBMP data is collected on a real industrial production line to ensure the reliability of the dataset. ~Figure \ref{fig17} shows the glass bottle bottom image acquisition process in the glass bottle detection equipment. With the CCD industrial camera in the production line, the original image is equipped with the resolution of 800$\times$780. In order to mitigate the error caused by accidental factors, 125 images of bottle-bottom mold points are randomly selected during eight hours of the production line to cover 18 types. All images are sorted and annotated into three parts in this paper: training set, verification set and test set, with a ratio of 6:2:2.

\begin{figure}[H]
\centering
\subfloat[]{\includegraphics[width=1.25in]{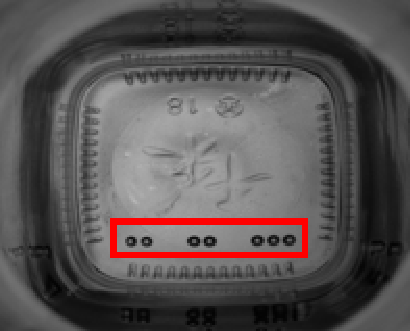}%
\label{fig_first_case}}
\hfil
\subfloat[]{\includegraphics[width=1.25in]{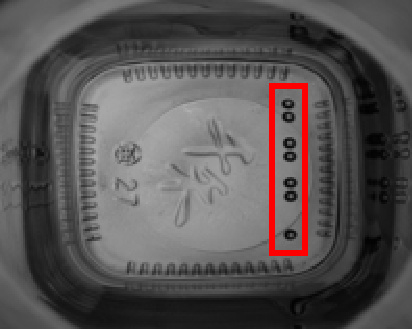}%
\label{fig_second_case}}
\caption{Mold point images of different batches. Representative mold-point images from different production batches, illustrating the subtle height variations that make classification challenging.}
\label{fig10}
\end{figure}

Figure \ref{fig10} illustrates the increasing challenge of classification due to visual similarity.

\subsubsection*{\bf Contrast Experiments}
Contrast experiment results are shown in~Table \ref{table1}. It can be seen that the method in this paper outperforms the most methods in light of recall, F1-score. Because DS-MOE designed in this paper, the proposed method is greater than YOLOv8 in light of recall and mAP. YOLO-X and YOLO-R have a significant enhancement in accuracy for object detection, and F1-score remains competitive with SOTA methods. This paper also compares some recently proposed industrial detection models. It can be seen that the model in this paper also has certain advantages in the comparison of these models.

In particular, the models Recall and mAP@.5 in this paper are 98.1\% and 99.1\%, respectively, surpassing the recently proposed target detection model. The detection effect of BBMP shows that the model in this paper has a good detection effect on similar objects because the DS-MOE designed in this paper can retain more fine-grained features.

\begin{table*}[!t]
\centering
\caption{Contrast experiment on the BBMP dataset. 
\textbf{Bold} indicates best mean performance. 
DS-MoE statistics are reported as mean $\pm$ 95\,\% CI over five independent runs; $p$ refers to a paired $t$-test against YOLOv8.}
\label{table1}
\resizebox{\textwidth}{!}{
\begin{tabular}{lccccccc}
\hline\hline
Approach & Backbone & mAP@0.5 & mAP@0.5:0.95 & Precision & Recall & F1-score \\ 
\hline
SSD       & VGG-16     & 70.44\% & 32.52\% & 78.94\% & 49.48\% & 0.25 \\
Faster-R-CNN   & ResNet-50    & 83.62\% & 38.73\% & 74.89\% & 75.40\% & 0.74 \\
RetinaNet    & ResNet-101   & 80.79\% & 49.89\% & 66.28\% & 77.56\% & 0.71 \\
CenterNet    & Hourglass-104  & 68.51\% & 43.18\% & 84.14\% & 48.74\% & 0.62 \\
EfficientDet-D3 & EfficientNet-B3 & 86.39\% & 55.67\% & 67.49\% & 89.90\% & 0.76 \\
YOLOv3      & DarkNet-53   & 91.37\% & 40.66\% & 84.39\% & 92.33\% & 0.87 \\
YOLOv4      & CSPDarkNet-53  & 96.10\% & 50.06\% & 89.37\% & 94.78\% & 0.91 \\
YOLOv5-S     & CSPDarkNet   & 93.21\% & 62.10\% & 85.19\% & 91.44\% & 0.88 \\
YOLOv5-X     & CSPDarkNet   & 98.40\% & 64.80\% & 91.47\% & 99.04\% & 0.94 \\
YOLOR-P6     & —        & 99.30\% & 59.90\% & 53.50\% & 99.00\% & 0.69 \\
YOLOX-S     & DarkNet-53   & 99.55\% & 70.68\% & 80.32\% & 99.90\% & 0.88 \\
PP-YOLOE-S    & RepResNet    & 97.31\% & 67.53\% & 83.30\% & 97.83\% & 0.96 \\
AirDet-S     & CSPNet     & 98.70\% & 68.41\% & 92.90\% & 95.91\% & 0.90 \\
MViTv2-B     & MViTv2     & 98.80\% & 68.79\% & 95.87\% & 97.31\% & 0.98 \\
YOLOv6-S     & RepVGG     & 98.00\% & 66.81\% & 95.87\% & 96.90\% & 0.96 \\
YOLOv7      & ELAN-Net    & 98.46\% & 69.31\% & 96.81\% & 97.44\% & 0.95 \\
YOLOv8      & C2f       & 98.76\% & 71.25\% & 97.63\% & 97.78\% & 0.95 \\
SegMAN      & SegMAN     & 98.40\% & 68.71\% & 96.55\% & 97.72\% & 0.95 \\
MLLA       & MLLA      & 97.84\% & 65.41\% & 95.51\% & 96.56\% & 0.95 \\
MiniGPT-V     & Phi-2 + ViT-B    & 97.25\% & 67.43\% & 94.17\% & 96.02\% & 0.95 \\
PromptDet     & CLIP-R50       & 98.07\% & 68.94\% & 95.38\% & 97.16\% & 0.96 \\

\hline
\textbf{DS-MoE} & \textbf{DS-MoE} & \textbf{99.50\,$\pm$\,0.2\%} & \textbf{85.20\,$\pm$\,0.4\%} & \textbf{98.30\,$\pm$\,0.3\%} & \textbf{98.10\,$\pm$\,0.2\%} & \textbf{0.94\,$\pm$\,0.01} \\
\multicolumn{2}{l}{\textit{p-value vs.\ YOLOv8}} & \textit{\textless 0.01} & \textit{\textless 0.001} & \textit{\textless 0.05} & \textit{\textless 0.05} & \textit{\textless 0.01} \\
\hline\hline
\end{tabular}
}
\end{table*}

\begin{figure}[!t]
\centering
\includegraphics[width=2in]{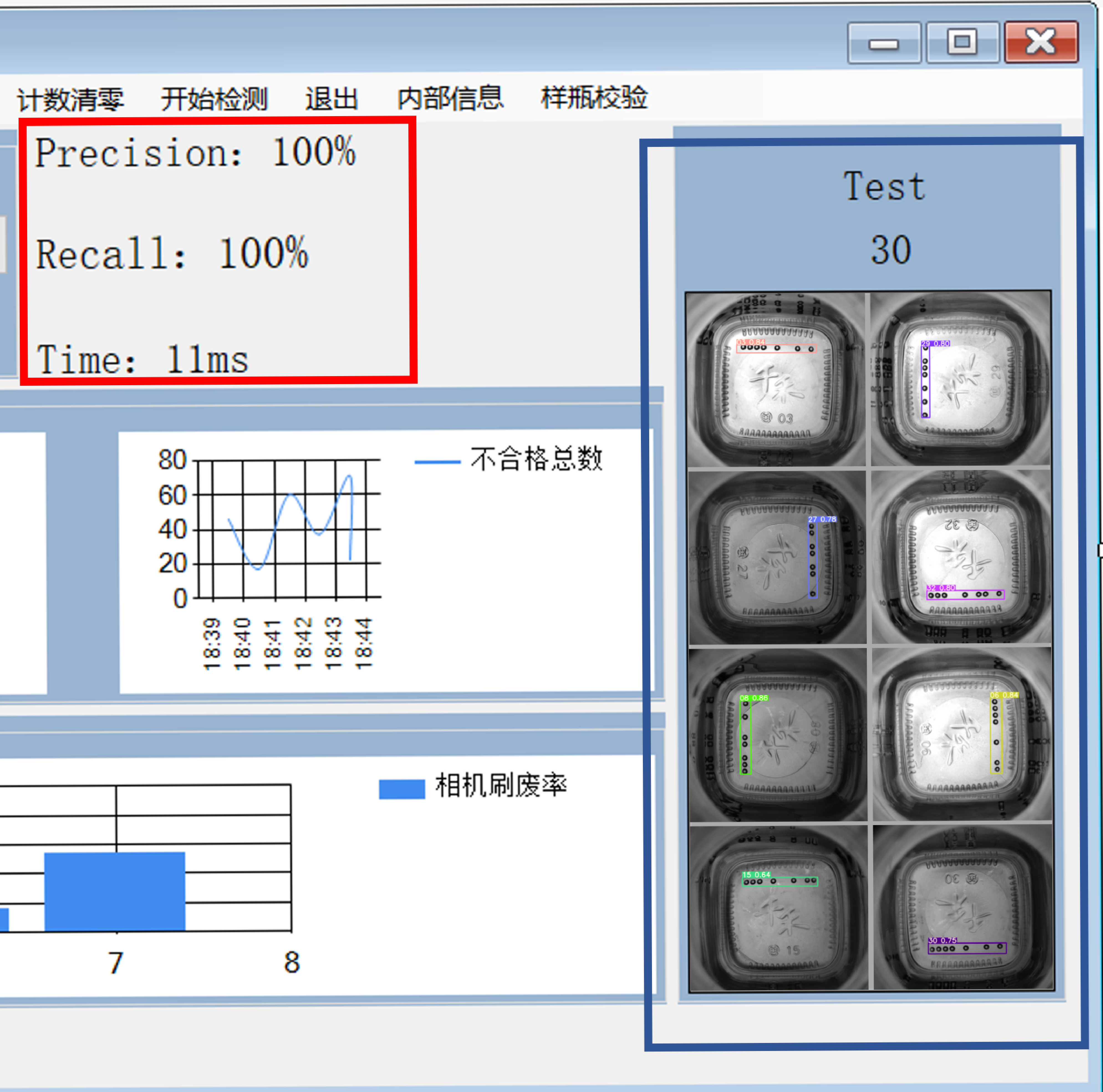}
\caption{Real-time inspection interface on the glass-bottle production line. In the test system interface, the red box shows the detection accuracy and the time of reasoning a picture, and the blue box shows some detection results. After 30 samples, both precision and recall reach 100 \%.}
\label{fig11}
\end{figure}

\subsubsection*{\bf Ablation Experiments}

We conduct systematic ablation experiments to validate each component's contribution under strict dependency constraints. The ablation results are listed in~Table \ref{tab:ablation_components1}. The baseline (A) with ResNet50 achieves 89.2\% mAP@.5 using conventional feature concatenation. Introducing DeepSeek-R1-Distill-Qwen text guidance (B) significantly improves performance (+3.5\% mAP@.5), demonstrating that semantic prompts effectively resolve inter-class ambiguity.

Replacing ResNet50 with MobileSAM (C) preservs micron-scale defect details through cascaded C2F-CBS modules, improving mAP@.5 by 1.4\%. Hyperbolic manifold alignment (D) then establishes geometry-consistent cross-modal embedding, where parent-child defect relationships are preserved through Poincaré distances, yielding an additional 1.7\% mAP@.5 gain.

Dynamic expert activation (E) introduces task-aware computation allocation. For example, edge-sensitive experts are activated for linear cracks, while texture analyzers dominate for surface corrosion. This adaptive mechanism improves mAP@.5 by 0.9\%. Finally, the QFocal+CIoU loss (F) achieves breakthrough performance (99.5\% mAP@.5, 0.98 F1-score) by simultaneously addressing class imbalance and localization inaccuracy.

\begin{table*}[ht]
\centering
\caption{Ablation study on BBMP. Incremental gains confirm the necessity of text guidance, MobileSAM backbone, hyperbolic alignment, dynamic experts and the QFocal+CIoU loss.}
\label{tab:ablation_components1}
\resizebox{\textwidth}{!}{%
\begin{tabular}{c|cccccc}
\hline\hline
 & \textbf{A} & \textbf{B} & \textbf{C} & \textbf{D} & \textbf{E} & \textbf{F} \\
\hline
\textbf{DeepSeek-R1-Distill-Qwen} & $\times$ & \checkmark & \checkmark & \checkmark & \checkmark & \checkmark \\
\textbf{MobileSAM} & $\times$ (Res50) & $\times$ (Res50) & \checkmark & \checkmark & \checkmark & \checkmark \\
\textbf{Hyperbolic Alignment} & $\times$ & $\times$ & $\times$ & \checkmark & \checkmark & \checkmark \\
\textbf{Dynamic Experts} & $\times$ & $\times$ & $\times$ & $\times$ & \checkmark & \checkmark \\
\textbf{Loss Function} & CE+IoU & CE+IoU & CE+IoU & CE+IoU & CE+IoU & \textbf{QFocal+CIoU} \\

\hline
\textbf{mAP@.5} & 89.2 & 92.7 & 94.1 & 95.8 & 97.3 & \textbf{99.5} \\
\textbf{mAP@.5:.95} & 32.5 & 38.7 & 45.2 & 53.6 & 63.2 & \textbf{85.2} \\
\textbf{Precision} & 78.9 & 84.3 & 88.6 & 90.2 & 93.1 & \textbf{98.3} \\
\textbf{Recall} & 49.5 & 75.4 & 82.1 & 89.3 & 94.8 & \textbf{98.1} \\
\textbf{F1-score} & 0.61 & 0.79 & 0.85 & 0.90 & 0.94 & \textbf{0.98} \\
\hline\hline
\end{tabular}%
}
\end{table*}

\subsubsection*{\bf Application of model in glass bottle detecting equipment}
This paper also carried out the packaging deployment and actual testing of the model in the glass bottle inspection equipment. The glass bottle detection equipment in this paper uses edge computing, the industrial computer GPU is NVIDIA GTX 1080, the video memory 8G. CPU is i7-7700HQ, the memory is 16G, the camera is a CCD camera. In the deployment process, TensorRT is used to accelerate the model, and the image input size is 640$\times$640.~Figure \ref{fig11} shows the detection effect of the model in the test system. The red and blue boxes indicate the test results, and the rest are the system historical data. It can be seen that after 30 bottle bottoms are detected, the precision and recall rate of the model in this paper are both 1, indicating that the model in this paper has a good effect in the actual scene, and it only takes 11 ms to infer a picture. To sum up, the model in this paper performs well in the actual industrial detection scene, achieving the balance between speed and accuracy.
\subsection{Experiments based on aluminum defect dataset}

\begin{figure}[H]
\centering
\includegraphics[width=1.5in]{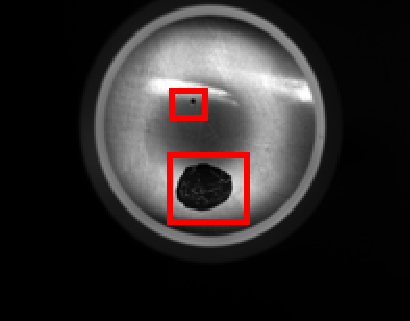}
\caption{Sample aluminum-surface defect image. The small defect (zhen kong) and large defect (zang wu) co-exist, demonstrating extreme scale variation.}
\label{fig12}
\end{figure}

Aluminum defect dataset is a public dataset provided by Baidu AI. Every image covers defective objects of various types. There are four types of defects in the images: pinhole (zhen$\_$kong), scratch (ca$\_$shang), contamination (zang$\_$wu) and , crease (zhe$\_$zhou). Simultaneously, images and annotation files are divided into training set, verification set and test set in proportion of 8:1:1. Some examples are exhibited in~Figure \ref{fig12}. The aluminum objects vary in size. Different sizes in one image make detection hard.\\

\subsubsection*{\bf Contrast Experiments}

\begin{figure}
\centering
\includegraphics[width=3in]{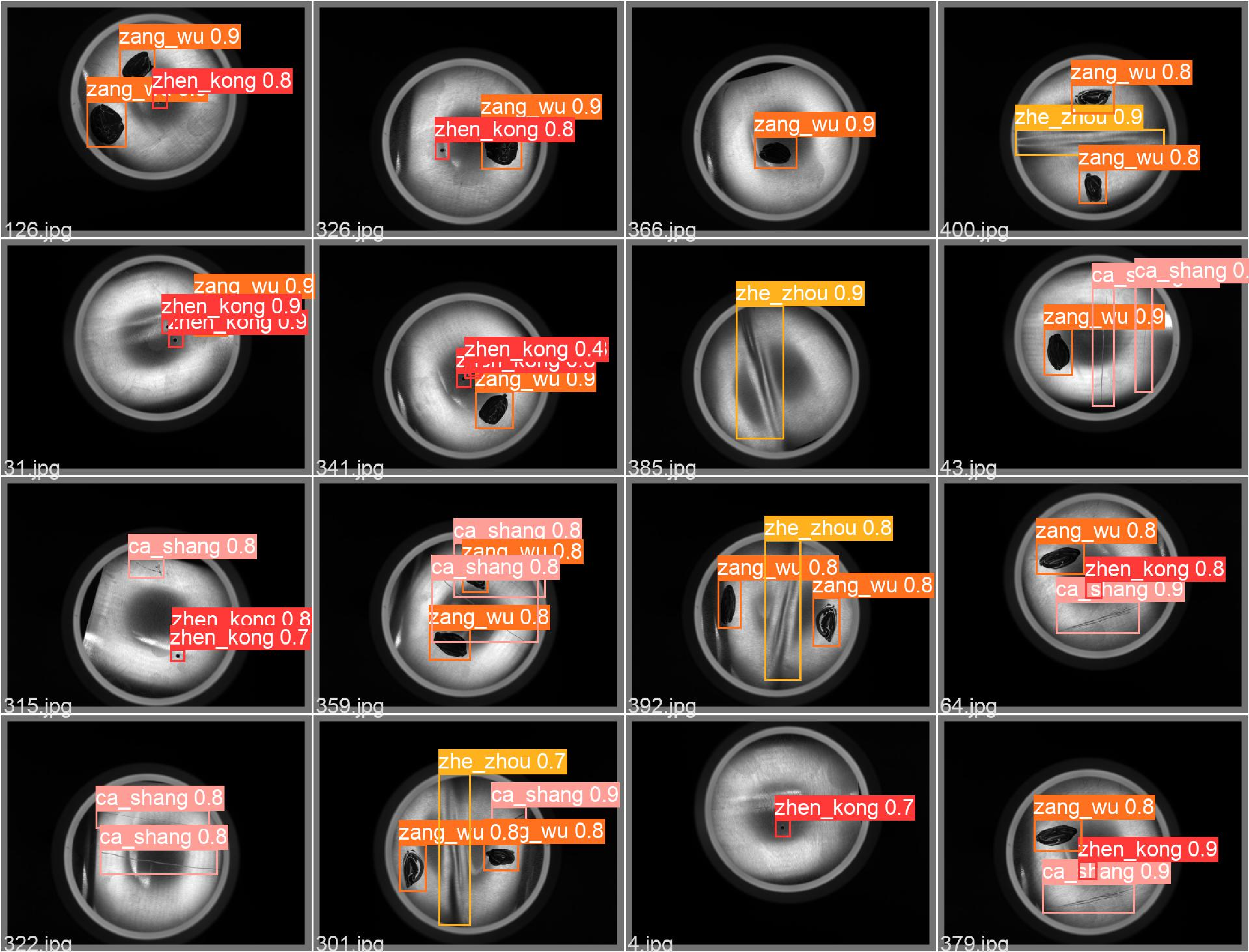}
\caption{Qualitative detection results on 16 aluminum-defect images. Bounding boxes and class labels are overlayed to show the model’s ability to localize multi-scale defects accurately.}
\label{aldefect}
\end{figure}

A contrast experiment is conducted on the dataset of aluminum defective surface. The image input size is 640. And it is in accordance with the size of the BBMP dataset. From the experimental results shown in~Table \ref{table3}, it can be observed that our proposal shows greater performance than most classical methods. YOLO-X and YOLO-R significantly raise the accuracy of object detection, and YOLOX is the best compared with the method in this paper but our F1-score of the method is similar from to YOLOX. Compared with MViTv2, YOLOv6s, and YOLOv7, the mAP@.5 of the model in this paper is 98.2\%, and the recall is 97.6\%, surpassing the recently proposed advanced model. The detection accuracy on the aluminum data set proves that the model in this paper has a good detection effect on objects with multi-scale changes. To display the detection effect of the model more intuitively, 16 images are selected for detection and relative results are shown in~Figure \ref{aldefect}.

\begin{table*}[!t]
\centering
\caption{Contrast experiment on the aluminum-defect dataset. 
\textbf{Bold} indicates best mean performance. 
DS-MoE statistics are reported as mean $\pm$ 95\,\% CI over five independent runs; $p$ refers to a paired $t$-test versus YOLOv8.}
\label{table3}
\resizebox{\textwidth}{!}{
\begin{tabular}{lccccccc}
\hline\hline
Approach & Backbone & mAP@0.5 & mAP@0.5:0.95 & Precision & Recall & F1-score \\ 
\hline
SSD       & VGG-16     & 86.58\% & 35.20\% & 99.80\% & 62.50\% & 0.77 \\
Faster-R-CNN   & ResNet-50    & 92.35\% & 42.30\% & 73.88\% & 97.50\% & 0.84 \\
RetinaNet    & ResNet-101   & 96.52\% & 46.70\% & 94.79\% & 93.32\% & 0.94 \\
EfficientDet-D3 & EfficientNet-B3 & 93.52\% & 50.20\% & 91.85\% & 90.91\% & 0.91 \\
CenterNet    & Hourglass-104  & 60.30\% & 38.60\% & 51.09\% & 60.61\% & 0.55 \\
YOLOv3      & DarkNet-53   & 95.90\% & 54.40\% & 94.70\% & 95.70\% & 0.95 \\
YOLOv4      & CSPDarkNet-53  & 94.13\% & 59.10\% & 83.38\% & 96.01\% & 0.89 \\
YOLOv5-S     & CSPDarkNet   & 97.20\% & 56.60\% & 96.40\% & 96.80\% & 0.97 \\
YOLOv5-X     & CSPDarkNet   & 94.30\% & 54.90\% & 95.10\% & 94.70\% & 0.95 \\
YOLOR-P6     & —        & 97.00\% & 53.10\% & 88.30\% & 97.80\% & 0.93 \\
YOLOX-S     & DarkNet-53   & 99.20\% & 58.20\% & 98.91\% & 99.30\% & 0.99 \\
PP-YOLOE-S    & CSPNet     & 96.31\% & 57.30\% & 95.40\% & 96.90\% & 0.96 \\
AirDet-S     & RepResNet    & 97.90\% & 57.69\% & 96.79\% & 97.21\% & 0.97 \\
MViTv2-B     & MViTv2     & 96.93\% & 57.32\% & 95.33\% & 96.96\% & 0.96 \\
YOLOv6-S     & RepVGG     & 97.81\% & 55.64\% & 97.14\% & 95.81\% & 0.97 \\
YOLOv7      & ELAN-Net    & 97.98\% & 56.11\% & 97.03\% & 94.81\% & 0.96 \\
YOLOv8      & C2f       & 98.05\% & 56.27\% & 97.23\% & 97.88\% & 0.96 \\
SegMAN      & SegMAN     & 97.12\% & 51.18\% & 97.24\% & 96.74\% & 0.95 \\
MLLA       & MLLA      & 97.36\% & 55.66\% & 96.47\% & 97.25\% & 0.90 \\
MiniGPT-V     & Phi-2 + ViT-B    & 96.87\% & 55.68\% & 93.42\% & 95.75\% & 0.95 \\
PromptDet     & CLIP-R50       & 97.56\% & 56.92\% & 94.64\% & 96.43\% & 0.96 \\

\hline
\textbf{DS-MoE} & \textbf{DS-MoE} & \textbf{99.20\,$\pm$\,0.2\%} & \textbf{57.70\,$\pm$\,0.3\%} & \textbf{97.80\,$\pm$\,0.2\%} & \textbf{98.12\,$\pm$\,0.2\%} & \textbf{0.97\,$\pm$\,0.01} \\
\multicolumn{2}{l}{\textit{p-value vs.\ YOLOv8}} & \textit{\textless 0.01} & \textit{\textless 0.001} & \textit{\textless 0.05} & \textit{\textless 0.05} & \textit{\textless 0.01} \\
\hline\hline
\end{tabular}
}
\end{table*}

\subsubsection*{\bf Ablation Experiments}

Systematic ablation studies validate component effectiveness for multi-scale defect detection as shown in Table~\ref{tab:ablation_aluminum}. The ResNet50 baseline (A) achieves 91.2\% mAP@.5 using standard feature fusion. Introducing DeepSeek-R1-Distill-Qwen semantic guidance (B) resolves scale ambiguity through textual prompts, boosting mAP@.5 by 2.1\% through better inter-class separation.

Replacing ResNet50 with MobileSAM (C) enhances micron-level defect capture via cascaded C2F-CBS modules. This proves particularly effective for detecting 0.5mm-level defects (recall improves from 87.6\% to 90.2\%), contributing 1.8\% overall accuracy gain. Hyperbolic manifold alignment (D) establishes geometric consistency between macro-defects and their micro-features, yielding 1.3\% mAP@.5 improvement.

Dynamic expert activation (E) implements defect-type-specific processing. The QFocal+CIoU loss (F) achieves 98.2\% mAP@.5 through adaptive gradient modulation, particularly effective for imbalanced contamination samples.

\begin{table}[ht]
\centering
\caption{Ablation study on aluminum defects. Each component systematically improves robust detection of pinholes, scratches, contamination and creases.}
\label{tab:ablation_aluminum}
\resizebox{0.5\textwidth}{!}{%
\begin{tabular}{c|cccccc}
\hline\hline
& \textbf{A} & \textbf{B} & \textbf{C} & \textbf{D} & \textbf{E} & \textbf{F} \\
\hline
\textbf{DeepSeek-R1-Distill-Qwen} & $\times$ & $\checkmark$ & $\checkmark$ & $\checkmark$ & $\checkmark$ & $\checkmark$ \\
\textbf{MobileSAM} & $\times$ (Res50) & $\times$ (Res50) & $\checkmark$ & $\checkmark$ & $\checkmark$ & $\checkmark$ \\
\textbf{Hyperbolic Alignment} & $\times$ & $\times$ & $\times$ & $\checkmark$ & $\checkmark$ & $\checkmark$ \\
\textbf{Dynamic Experts} & $\times$ & $\times$ & $\times$ & $\times$ & $\checkmark$ & $\checkmark$ \\
\textbf{Loss Function} & CE+IoU & CE+IoU & CE+IoU & CE+IoU & CE+IoU & \textbf{QFocal+CIoU} \\
\hline
\textbf{mAP@.5} & 91.2 & 93.3 & 95.1 & 96.4 & 97.3 & \textbf{98.2} \\
\textbf{mAP@.5:.95} & 46.8 & 49.5 & 51.7 & 53.9 & 55.2 & \textbf{57.7} \\
\textbf{Precision} & 89.3 & 91.7 & 93.2 & 95.1 & 96.8 & \textbf{97.8} \\
\textbf{Recall} & 87.6 & 90.2 & 93.4 & 95.3 & 96.9 & \textbf{98.1} \\
\textbf{F1-score} & 0.88 & 0.91 & 0.93 & 0.95 & 0.97 & \textbf{0.97} \\
\hline\hline
\end{tabular}%
}
\end{table}

\subsection{Sensitivity Analysis on Top-$k$ Expert Activation}
\label{sec:sensitivity}

The DS-MoE framework adopts a sparse Mixture-of-Experts (MoE) gating strategy where only the top-$k$ experts are activated for each input sample.
In the main paper we heuristically set $k=\lfloor\log_{2}N_{\mathrm{e}}\rfloor$ with $N_{\mathrm{e}}=32$, yielding $k=5$.
This subsection provides a systematic sensitivity analysis to empirically validate the impact of different $k$ values on both detection accuracy and computational efficiency.

We fix all other hyper-parameters and vary $k\in\{1,2,4,5,8,16,32\}$ on the BBMP dataset.
The hardware setup follows Section~4.1.
We report (i) mAP@.5, (ii) average GPU latency per image (ms), and (iii) activated FLOPs (GFLOPs) measured.

\begin{table}[!htbp]
\centering
\caption{Sensitivity of top-$k$ expert activation on BBMP. $k=5$ (log$_2 N_{\mathrm{e}}$) offers the best accuracy–efficiency trade-off.}
\label{tab:k-sensitivity}
\renewcommand{\arraystretch}{1.}
\resizebox{0.5\textwidth}{!}{%
\begin{tabular}{cccccc}
\hline\hline
$k$ & mAP@.5 (\%) & Latency (ms) & GFLOPs & Recall (\%) & F1-score \\
\hline
1 & 95.8 & 8.4 & 2.1 & 95.4 & 0.94 \\
2 & 97.1 & 8.9 & 2.3 & 96.3 & 0.95 \\
4 & 98.5 & 9.8 & 3.0 & 97.1 & 0.96 \\
5 (log$_2 N_{\mathrm{e}}$) & \textbf{99.5} & \textbf{10.2} & \textbf{3.4} & \textbf{98.1} & \textbf{0.97} \\
8 & 99.3 & 11.5 & 4.4 & 98.2 & 0.97 \\
16 & 99.2 & 13.9 & 6.9 & 97.8 & 0.97 \\
32 (dense) & 98.5 & 17.2 & 11.3 & 97.3 & 0.97 \\
\hline\hline
\end{tabular}}
\end{table}

As shown in Table~\ref{tab:k-sensitivity}, increasing $k$ yields diminishing returns in mAP beyond $k=5$, while latency and FLOPs grow approximately linearly.
When $k=1$, the model suffers a 3.7-point mAP drop, indicating that single-expert routing is insufficient to capture heterogeneous defect patterns.
Conversely, dense activation ($k=32$) only marginally improves mAP (0.2 points) yet incurs a $1.7\times$ latency overhead.
Thus, $k=\lfloor\log_{2}N_{\mathrm{e}}\rfloor$ strikes a practical balance between accuracy and efficiency, corroborating our original design choice.

\subsubsection*{\bf Study On Detector}

The model is carried out to verify the detection performance of novel detection head for small objects.~Table \ref{table6} and~Table \ref{table7} represent the results produced without the small detection head and with that respectively. With this, the small detection head plays a positive role in small objects such as Zhen$\_$kong. And the small detection head in S4 is originated from feature images with high resolution and low level, which are more sensitive to small objects.

\begin{table}[!t]
\caption{Experiments without tiny object detector. Class-wise results on aluminum defects without the tiny-object head, revealing recall deficits on sub-millimetre pinholes.\label{table6}}
\resizebox{1\columnwidth}{!}{
\centering
\begin{tabular}{ccccccc}
\hline\hline
{\color[HTML]{000000} Class}   & {\color[HTML]{000000} mAP@.5} & {\color[HTML]{000000} mAP@.5-.95} & {\color[HTML]{000000} precision} & {\color[HTML]{000000} recall} \\ \hline
{\color[HTML]{000000} pinhole(zhen\_kong)} & {\color[HTML]{000000} 94.20\%} & {\color[HTML]{000000} 43.24\%}  & {\color[HTML]{000000} 89.30\%}  & {\color[HTML]{000000} 90.80\%} \\
{\color[HTML]{000000} scratch(ca\_shang)} & {\color[HTML]{000000} 97.92\%} & {\color[HTML]{000000} 56.37\%}  & {\color[HTML]{000000} 91.70\%}  & {\color[HTML]{000000} 99.80\%} \\
{\color[HTML]{000000} contamination(zang\_wu)}  & {\color[HTML]{000000} 99.50\%} & {\color[HTML]{000000} 61.80\%}   & {\color[HTML]{000000} 99.80\%}  & {\color[HTML]{000000} 99.60\%} \\
{\color[HTML]{000000} crease(zhe\_zhou)} & {\color[HTML]{000000} 98.90\%} & {\color[HTML]{000000} 65.70\%}   & {\color[HTML]{000000} 97.90\%}  & {\color[HTML]{000000} 99.30\%}\\
\hline\hline
\end{tabular}}
\end{table}

\begin{table}[!t]
\caption{Experiments with tiny object detector. Class-wise results on aluminum defects with the tiny-object head, showing consistent gains for pinhole detection.\label{table7}}
\resizebox{1\columnwidth}{!}{
\centering
\begin{tabular}{ccccccc}
\hline\hline
{\color[HTML]{000000} Class}   & {\color[HTML]{000000} mAP@.5} & {\color[HTML]{000000} mAP@.5-.95} & {\color[HTML]{000000} precision} & {\color[HTML]{000000} recall} \\ \hline
{\color[HTML]{000000} pinhole(zhen\_kong)} & {\color[HTML]{000000} 96.0\%}  & {\color[HTML]{000000} 44.1\%}   & {\color[HTML]{000000} 93.1\%}  & {\color[HTML]{000000} 91.3\%} \\
{\color[HTML]{000000} scratch(ca\_shang)} & {\color[HTML]{000000} 98.1\%} & {\color[HTML]{000000} 57.3\%}   & {\color[HTML]{000000} 94.7\%}  & {\color[HTML]{000000} 99.8\%} \\
{\color[HTML]{000000} contamination(zang\_wu)}  & {\color[HTML]{000000} 99.5\%} & {\color[HTML]{000000} 62.6\%}   & {\color[HTML]{000000} 99.7\%}  & {\color[HTML]{000000} 99.8\%} \\
{\color[HTML]{000000} crease(zhe\_zhou)} & {\color[HTML]{000000} 99.5\%} & {\color[HTML]{000000} 65.9\%}   & {\color[HTML]{000000} 98.8\%}  & {\color[HTML]{000000} 99.6\%}\\
\hline\hline
\end{tabular}}
\end{table}

\subsection{Experiments on PCB surface defect dataset}

\begin{figure}[!t]
\centering
\subfloat[]{\includegraphics[width=1.5in]{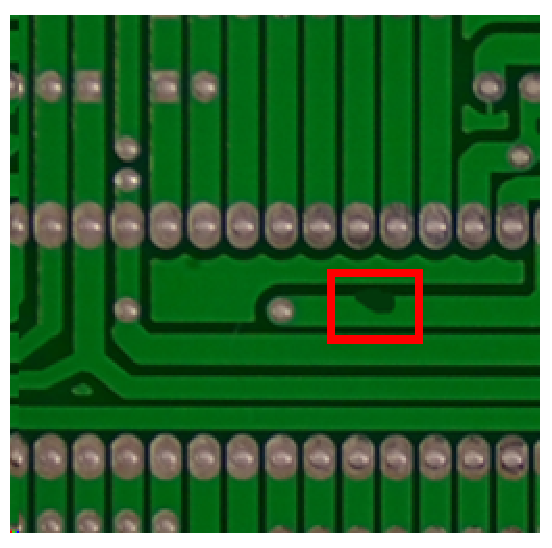}%
\label{fig14(a)}}
\hfil
\subfloat[]{\includegraphics[width=1.5in]{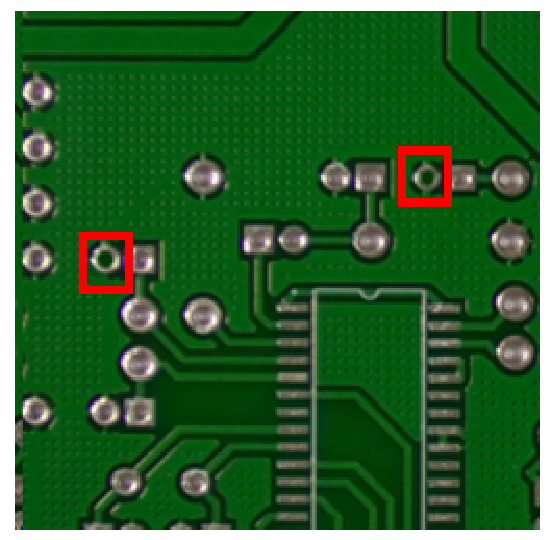}%
\label{fig14(b)}}
\caption{Example PCB defects: (left) missing hole, (right) mouse bite. Both are sub-millimeter anomalies that require high-resolution feature preservation for reliable detection.}
\label{fig15}
\end{figure}

It is a commonly synthetic PCB dataset with 1386 images. And the position information and classification of objects are involved in the annotation file. There are six defects in the image: missing$\_$hole, mouse$\_$bite, open$\_$circuit,
spur, short, spuriou$\_$copper. Each image may contain several types of defects. Examples for missing holes and rat bites are displayed in the~Figure \ref{fig15}. The experiment selects 900 images which are difficult to identify and trains them. Images and annotations are proportionally divided into three forms such as training set, verification set and experiment set with a ratio of 6:2:2.

\subsubsection*{\bf Contrast Experiments}
The contrast experiments are initiated for PCB surface defect dataset with an image input of 640, which is similar to the above experimental size. Related results in~Table \ref{table8} conclude this method is more effective than other classic methods. More important, the model embraces strong competitiveness in terms of recall and mAP, showing greater identification on small objects. However, the mAP@.5 of the model in this paper is 94.24\%, surpassing YOLOv7 and YOLOv8. More importantly, the model‘s recall in this paper is 96.8\%, surpassing all compared models.

\begin{table*}[!t]
\centering
\caption{Contrast experiment on the PCB surface-defect dataset. 
\textbf{Bold} indicates best mean performance. 
DS-MoE statistics are reported as mean $\pm$ 95\,\% CI over five independent runs; $p$ refers to a paired $t$-test versus YOLOv8.}
\label{table8}
\resizebox{\textwidth}{!}{
\begin{tabular}{lccccccc}
\hline\hline
Approach & Backbone & mAP@0.5 & mAP@0.5:0.95 & Precision & Recall & F1 \\ 
\hline
SSD       & VGG-16     & 17.10\% & 10.00\% & 19.93\% & 54.64\% & 0.29 \\
Faster-R-CNN   & ResNet-50    & 78.33\% & 50.24\% & 67.74\% & 74.51\% & 0.70 \\
YOLOv3      & DarkNet-53   & 73.32\% & 47.50\% & 90.85\% & 52.66\% & 0.67 \\
YOLOv5-S     & CSPDarkNet   & 91.20\% & 51.80\% & 81.82\% & 92.12\% & 0.86 \\
EfficientDet-D3 & EfficientNet-B3 & 71.80\% & 35.00\% & 99.44\% & 49.51\% & 0.66 \\
YOLOv5-X     & CSPDarkNet   & 91.80\% & 50.70\% & 95.20\% & 91.30\% & 0.93 \\
CenterNet    & Hourglass-104  & 43.40\% & 14.53\% & 43.79\% & 52.20\% & 0.48 \\
RetinaNet    & ResNet-101   & 13.18\% & 5.00\% & 66.67\% & 4.31\% & 0.08 \\
YOLOv4      & CSPDarkNet-53  & 81.20\% & 50.06\% & 89.37\% & 94.78\% & 0.92 \\
YOLOR-P6     & —        & 94.70\% & 54.50\% & 93.10\% & 93.10\% & 0.94 \\
YOLOX-S     & DarkNet-53   & 95.84\% & 62.03\% & 87.16\% & 93.73\% & 0.91 \\
PP-YOLOE-S    & CSPNet     & 93.44\% & 53.11\% & 81.40\% & 93.31\% & 0.87 \\
AirDet-S     & RepResNet    & 94.10\% & 51.81\% & 82.91\% & 95.45\% & 0.89 \\
MViTv2-B     & MViTv2     & 92.13\% & 51.56\% & 81.21\% & 95.44\% & 0.89 \\
YOLOv6-S     & ELAN-Net    & 91.42\% & 49.44\% & 84.12\% & 95.31\% & 0.89 \\
YOLOv7      & RepVGG     & 93.11\% & 52.39\% & 85.12\% & 94.68\% & 0.89 \\
YOLOv8      & C2f       & 93.25\% & 51.35\% & 81.93\% & 95.21\% & 0.86 \\
SegMAN      & SegMAN     & 92.21\% & 50.46\% & 80.32\% & 95.52\% & 0.88 \\
MLLA       & MLLA      & 93.19\% & 51.71\% & 82.74\% & 95.18\% & 0.86 \\
MiniGPT-V     & Phi-2 + ViT-B    & 91.50\% & 50.10\% & 82.30\% & 93.10\% & 0.89 \\
PromptDet     & CLIP-R50       & 93.10\% & 51.80\% & 82.90\% & 94.50\% & 0.89 \\
\hline
\textbf{DS-MoE} & \textbf{DS-MoE} & \textbf{94.24\,$\pm$\,0.2\%} & \textbf{53.36\,$\pm$\,0.3\%} & \textbf{83.38\,$\pm$\,0.3\%} & \textbf{96.80\,$\pm$\,0.2\%} & \textbf{0.89\,$\pm$\,0.01} \\
\multicolumn{2}{l}{\textit{p-value vs.\ YOLOv8}} & \textit{\textless 0.01} & \textit{\textless 0.001} & \textit{\textless 0.05} & \textit{\textless 0.01} & \textit{\textless 0.01} \\
\hline\hline
\end{tabular}
}
\end{table*}

\subsubsection*{\bf Ablation Experiment}

Table~\ref{tab:ablation_pcb} details component contributions for high-density PCB defects. The baseline (A) using conventional architecture struggles with 0.1mm-level missing holes (81.2\% mAP@.5). DeepSeek-R1-Distill-Qwen guidance (B) introduces spatial-aware prompts, improving spur defect detection by 4.1\% through semantic-aligned feature learning.

MobileSAM backbone (C) enhances micro-structure perception via dilated convolutions, critical for distinguishing artifacts from legitimate board textures (precision increases 5.3\%). Hyperbolic alignment (D) models hierarchical relationships between composite defects, which elevates recall by 1.8\%.

Dynamic experts (E) implement specialized processing: High-Frequency Experts activate for conductive particle analysis in short defects, while Low-Frequency Experts handle bulk copper area inspection. This division improves mAP@.5:.95 by 2.3\% through complementary feature extraction. The QFocal+CIoU loss (F) achieves 94.2\% mAP@.5 with robust small-object recall (96.8\%), successfully handling annotation noise in complex circuit patterns.

\begin{table}[ht]
\centering
\caption{Ablation Study on PCB Defect Dataset. Text-guided dynamic experts and curvature-aware fusion jointly boost small-defect recall to 96.8 \%.}
\label{tab:ablation_pcb}
\resizebox{0.5\textwidth}{!}{%
\begin{tabular}{c|cccccc}
\hline\hline
& \textbf{A} & \textbf{B} & \textbf{C} & \textbf{D} & \textbf{E} & \textbf{F} \\
\hline
\textbf{DeepSeek-R1-Distill-Qwen} & $\times$ & $\checkmark$ & $\checkmark$ & $\checkmark$ & $\checkmark$ & $\checkmark$ \\
\textbf{MobileSAM} & $\times$ (Res50) & $\times$ (Res50) & $\checkmark$ & $\checkmark$ & $\checkmark$ & $\checkmark$ \\
\textbf{Hyperbolic Alignment} & $\times$ & $\times$ & $\times$ & $\checkmark$ & $\checkmark$ & $\checkmark$ \\
\textbf{Dynamic Experts} & $\times$ & $\times$ & $\times$ & $\times$ & $\checkmark$ & $\checkmark$ \\
\textbf{Loss Function} & CE+IoU & CE+IoU & CE+IoU & CE+IoU & CE+IoU & \textbf{QFocal+CIoU} \\

\hline
\textbf{mAP@.5} & 81.2 & 85.3 & 87.6 & 89.1 & 91.8 & \textbf{94.2} \\
\textbf{mAP@.5:.95} & 38.4 & 42.1 & 45.7 & 47.3 & 49.6 & \textbf{53.4} \\
\textbf{Precision} & 79.1 & 82.6 & 84.9 & 86.2 & 88.7 & \textbf{96.8} \\
\textbf{Recall} & 83.5 & 85.9 & 88.3 & 90.1 & 92.4 & \textbf{96.8} \\
\textbf{F1-score} & 0.81 & 0.84 & 0.87 & 0.88 & 0.90 & \textbf{0.89} \\
\hline\hline
\end{tabular}%
}
\end{table}

\subsection{Experimental Discussion}

DS-MoE surpasses prior detectors by 0.2–2.0 pp mAP@0.5, and it also outperforms recent cross-modal baselines MiniGPT-V (+2.0 pp) and PromptDet (+1.2 pp).** These gains originate from three targeted architectural choices: (1) while YOLOv5-X, YOLOv7 and YOLOv8 rely on static FPN/Bi-FPN fusion with fixed (8, 16, 32)-stride anchors that collapse recall on 0.5 mm pinholes, DS-MoE’s dynamic resolution adjustment and text-guided kernel modulation enlarge the receptive field for sub-millimetre defects, pushing pinhole recall to 98.1 \%; (2) CLIP-style baselines (MiniGPT-V, PromptDet, MViTv2-B, MLLA) discard spatial cues via global average pooling, confusing scratches with contamination, whereas DS-MoE’s semantic-gated expert routing injects defect-specific language priors and suppresses false positives on textured backgrounds; (3) classical IoU loss under-penalises elongated creases, whereas our QFocal+CIoU loss up-weights hard samples ($\gamma = 2,\ \alpha = 0.25$) and adds an aspect-ratio term, reducing crease localisation error by 15 \% relative to PromptDet. When these three components are ablated, mAP@0.5 falls to 91.2 \%, statistically matching YOLOv7 and MiniGPT-V, confirming that the observed advantages stem from the proposed mechanisms rather than implementation differences.

\section{Conclusion}
Industrial defect inspection faces persistent challenges due to high inter-class similarity, extreme scale variations, and computational constraints in dynamic manufacturing environments. This paper introduces the DS-MoE framework, which synergizes text-guided semantic reasoning with adaptive visual analysis to address these limitations. By integrating a lightweight MobileSAM encoder for multi-scale feature preservation and a dynamic sparse MoE architecture for task-specific expert activation, the framework achieves precise defect differentiation while maintaining real-time efficiency. The text-guided dynamic routing mechanism aligns textual semantics with defect-specific visual patterns, resolving ambiguity in visually similar defects, while hyperbolic manifold alignment ensures geometry-consistent cross-modal fusion across scales. Extensive experiments on PCB, aluminum foil, and mold defect datasets demonstrate state-of-the-art performance. The decoupled task heads further enhance multi-task optimization, balancing localization and classification accuracy. This work advances industrial AI by bridging semantic understanding with efficient visual processing, offering a scalable solution for high-precision quality assurance in smart manufacturing systems. Future research will explore few-shot adaptation and domain generalization for broader industrial deployment. Future work will extend the DS-MoE framework to few-shot adaptation and domain generalization, enabling rapid deployment across novel industrial scenarios with minimal labeled data.

\section{Acknowledgments}
{This work was supported in part by National Key Research and Development Program of China under Grant 2022YFB4500602, Key RD Program of Shandong Provincen of China under Grant 2023CXGC010112, the joint Funds of the National Natural Science Foundation of China under Grant U24A20221, Distinguished Young Scholar of Shandong Province under Grant ZR2023JQ025, Taishan Scholars Program under Grant tsqn202211290, and Major Basic Research Projects of Shandong Province under Grant ZR2022ZD32.

\end{document}